\documentclass[11pt]{article}
\usepackage[a4paper, left=2.5cm, right=2.5cm, top=2cm, bottom=2cm]{geometry}
\usepackage{framed,multirow}
\usepackage{amssymb}
\usepackage{amsmath}
\usepackage{latexsym}
\usepackage{url}
\usepackage{hyperref}
\usepackage{enumitem}
\usepackage{diagbox}
\usepackage[table, dvipsnames]{xcolor}
\usepackage{soul}
\usepackage{graphicx}
\usepackage{wrapfig}
\usepackage[affil-it]{authblk}
\usepackage{makecell}
\usepackage{chngpage}

\definecolor{titlegray}{gray}{0.75}
\definecolor{lightgray}{gray}{0.75}
\definecolor{lightergray}{gray}{0.95}
\hypersetup{
    colorlinks,
    linkcolor={red!50!black},
    citecolor={blue!50!black},
    urlcolor={blue!50!black}
}
\DeclareMathOperator{\E}{\mathbb{E}}

\usepackage[compact]{titlesec}
\titlespacing{\section}{0pt}{2ex}{2ex}
\titlespacing{\subsection}{0pt}{2ex}{1ex}
\titlespacing{\subsubsection}{0pt}{1ex}{0.5ex}

\title{A Field of Experts Prior for Adapting Neural Networks at Test Time}
\author{Neerav Karani,
Georg Brunner,
Ertunc Erdil,
Simin Fei,
\\Kerem Tezcan,
Krishna Chaitanya,
Ender Konukoglu}
\affil{Biomedical Image Computing Group, Computer Vision Laboratory, ETH Z{\"u}rich
\thanks{Manuscript under review.
\\ All authors are with the Biomedical Image Computing Group at ETH Zurich, Switzerland (https://bmic.ee.ethz.ch/).
Corresponding author: Neerav Karani (nkarani@vision.ee.ethz.ch).}}
\date{\today}

\begin{document}
\maketitle
% ==========================================
% ABSTRACT
% ==========================================
\begin{abstract}
% robustness against distribution shifts
Supervised learning methods based on convolutional neural networks (CNNs) show promising performance in several medical image analysis tasks. Such performance, however, is marred in the presence of acquisition-related distribution shifts between training and test images.
% test time adaptation
Recently, it has been proposed to tackle this problem by fine-tuning trained CNNs for each test image. Such \emph{test-time-adaptation} (TTA) is a promising and practical strategy for improving robustness to distribution shifts as it requires neither data sharing between institutions nor annotating additional data.
% previous TTA methods
Previous TTA methods use a \emph{helper} model to increase similarity between outputs and/or features extracted from a test image with those of the training images. Such helpers, which are typically modeled using CNNs and trained in a self-supervised manner, can be task-specific and themselves vulnerable to distribution shifts in their inputs.
% proposed method
To overcome these problems, we propose to carry out TTA by matching the feature distributions of test and training images, as modelled by a field-of-experts (FoE) prior.
% To overcome these problems, we propose a field-of-experts (FoE) prior for TTA.
% Experts are filters learned in the task network
FoEs model complicated probability distributions as products of several simpler \textit{expert} distributions. We use the $1$D marginal distributions of a trained task CNN's features as the experts in the FoE model.
% Additional experts as PCA of patches from last layer features
Further, we carry out principal component analysis (PCA) of patches of the task CNN's features, and consider the distributions of the PCA loadings as additional experts.
% Experiments 
We extensively validate the method's efficacy on 5 MRI segmentation tasks (healthy tissues in 4 anatomical regions and lesion segmentation in 1 one anatomy), using data from 17 institutions, and on a MRI registration task, using data from 3 institutions.
% \editone{We find that the proposed FoE-based TTA provides comparable performance to state-of-the-art TTA methods, with two additional benefits: (1) improved stability (prevention of performance degradation during TTA, which is sporadically observed in other TTA methods) and (2) improved generality (performance improvement for lesion datasets as well as to applicability to multiple tasks).}
We find that the proposed FoE-based TTA is generically applicable in multiple tasks, and outperforms all previous TTA methods for lesion segmentation. For healthy tissue segmentation, the proposed method outperforms other task-agnostic TTA methods, but a previous TTA method which is specifically designed for segmentation performs the best for most of the tested datasets.
Our implementation is publicly available \href{https://github.com/neerakara/TTA-1D-Density-Matching}{here}.
%%%%
\end{abstract}
% ==========================================
% Introduction
% ==========================================
\section{Introduction}
\label{sec:introduction}

% =======================
% The distribution shift problem
% =======================
\subsection{The distribution shift problem}\label{subsec:ds_problem}
% \noindent \textbf{The distribution shift problem}:
Performance of convolutional neural networks (CNNs) trained using supervised learning degrades when the distributions of training and test samples differ.
%In supervised learning using CNNs, a training dataset of paired inputs and outputs, $\{x_i, y_i\} \sim \mathcal{P}_{tr}(X,Y)$, is used to learn the parameters of a CNN, $S_{\theta}(x)$, using empirical risk minimization (ERM) \cite{vapnik1992principles}\href{http://koza.if.uj.edu.pl/~krzemien/machine_learning2021/materials/vapnik-principles-of-risk-minimization-for-learning-theory.pdf}{$^{\uparrow}$}: $\theta^* = argmin_{\theta} E_{\mathcal{P}_{tr}(X,Y)} \: L(y, S_{\theta}(x))$.
%The optimal parameters, $\theta^*$, provide high prediction accuracy for test inputs from the training distribution, $\bar{x} \sim \mathcal{P}_{tr}$, but reduced prediction accuracy for test inputs from a different distribution, $\bar{x} \sim \mathcal{P}_{ts}$~\cite{yan2020mri}\href{https://pubs.rsna.org/doi/pdf/10.1148/ryai.2020190195}{$^{\uparrow}$},~\cite{zhang2021empirical}\href{https://dl.acm.org/doi/pdf/10.1145/3450439.3451878}{$^{\uparrow}$}.
This is known as the distribution shift (DS) problem\footnote{We use the acronym DS to refer to the singular 'distribution shift' as well as the plural 'distribution shifts', and call on the reader to infer the form based on the context.}.
Several types of DS are pertinent in medical imaging~\cite{castro2020causality}.
% \href{https://www.nature.com/articles/s41467-020-17478-w.pdf}{$^{\uparrow}$}.
We consider acquisition-related DS - that is, DS caused by variations in scanners and acquisition protocol parameters.
Such shifts are pervasive in clinical practice; thus, tackling them suitably is crucial for large-scale adoption of deep learning methods.

% =======================
% Categories of methods to tackle the DS problem
% =======================
% In the DS literature, the training distribution is commonly referred to as the source domain (SD) and the test distribution as the target domain (TD).
\subsection{Categories of methods to tackle the DS problem}\label{subsec:ml_settings}
% \vspace{5pt} \noindent \textbf{Categories of methods to tackle the DS problem}:
Due to its high practical relevance, the DS problem has attracted substantial attention in the research community.
%The DS problem is well-recognized as an important challenge to clinical translation of automated image analysis methods, and has attracted substantial attention in the research community in recent years.
In decreasing order of dependence on data from the test distribution, methods in the DS literature can be broadly categorized into the following groups:
transfer learning (TL)~\cite{van2014transfer},
% \href{https://ieeexplore.ieee.org/stamp/stamp.jsp?arnumber=6945865}{$^\uparrow$},
\cite{tajbakhsh2016convolutional},
% \href{https://ieeexplore.ieee.org/stamp/stamp.jsp?arnumber=7426826}{$^\uparrow$},
\cite{karani2018lifelong},
% \href{https://arxiv.org/pdf/1805.10170.pdf}{$^\uparrow$}
unsupervised domain adaptation (UDA)~\cite{kamnitsas2017unsupervised},
% \href{https://arxiv.org/pdf/1612.08894.pdf}{$^\uparrow$},
\cite{huo2018synseg},
% \href{https://ieeexplore.ieee.org/stamp/stamp.jsp?arnumber=8494797}{$^\uparrow$}
and domain generalization (DG)~\cite{dou2019domain},
% \href{https://proceedings.neurips.cc/paper/2019/file/2974788b53f73e7950e8aa49f3a306db-Paper.pdf}{$^\uparrow$},
\cite{zhang2019generalizing},
% \href{https://ieeexplore.ieee.org/stamp/stamp.jsp?arnumber=8995481}{$^\uparrow$},
\cite{billot2021synthseg}.
% \href{https://arxiv.org/pdf/2107.09559.pdf}{$^\uparrow$}.
% Among these three settings, DG is the most appealing as, contrary to the other settings, it does not require sharing data between institutions, annotating additional images nor lengthy training procedures.
Among these three settings, DG is the most appealing - contrary to other settings, it does not require sharing data between institutions or annotating additional images.
% due to the difficulty in annotating medical images as well as sharing them across institutions.
% Recently, it has been shown that the performance of CNNs trained with DG techniques can be further improved by adapting them using unlabelled image(s) from the test distribution.
% This bring us to the settings of source-free domain adaptation (SFDA)~\cite{bateson2021sourcefree}\href{https://arxiv.org/pdf/2108.03152.pdf}{$^\uparrow$} and test-time adaptation (TTA)~\cite{karani2020test}\href{https://www.sciencedirect.com/science/article/pii/S1361841520302711}{$^\uparrow$}~\cite{he2021autoencoder}\href{https://www.sciencedirect.com/science/article/pii/S1361841521001821}{$^\uparrow$}.
Recently, two new settings have been proposed - source-free domain adaptation (SFDA)
% \cite{bateson2021sourcefree}\href{https://arxiv.org/pdf/2108.03152.pdf}{$^\uparrow$}
\cite{bateson2020source}
% \href{https://link.springer.com/chapter/10.1007/978-3-030-59710-8_48}{$^{\uparrow}$}
and test-time adaptation (TTA)
\cite{karani2020test},
% \href{https://www.sciencedirect.com/science/article/pii/S1361841520302711}{$^\uparrow$},
\cite{he2021autoencoder}.
% \href{https://www.sciencedirect.com/science/article/pii/S1361841521001821}{$^\uparrow$}.
Here, the performance of CNNs trained with DG techniques is further improved by adapting them using unlabelled image(s) from the test distribution. Importantly, the adaptation in TTA or SFDA is done without access to data from the training distribution. Due to these advantages, we pose our work in the TTA setting.

% =======================
% Test-time adaptation
% =======================
%TTA~\cite{karani2020test}\href{https://www.sciencedirect.com/science/article/pii/S1361841520302711}{$^{\uparrow}$},~\cite{he2021autoencoder}\href{https://www.sciencedirect.com/science/article/pii/S1361841521001821}{$^{\uparrow}$} is a recent approach for improving the performance of deep NNs in the face of such DS.
\subsection{Test-time adaptation}\label{subsec:tta}
% \vspace{5pt} \noindent \textbf{Test-time adaptation}:
In TTA, the parameters of a previously trained CNN are adapted for each test image.
The subset of the parameters that get adapted per test image is a design choice.
Noting that acquisition-related DS manifest as contrast variations, one approach is to design the CNN as a concatenation of a \emph{shallow, image-specific contrast normalization CNN}, $z = N_\phi(x)$, followed by a deep task CNN that is shared by all training and test images, $y = S_\theta(z)$.
Here, $x$ is the input image, $z$ is the \emph{normalized} image, and $y$ is the output (e.g. segmentation, deformation field, enhanced image).
The image-specific parameters, $\phi$, are adapted by requiring adherence to a prior model, $H_\psi$, either in the output space \cite{karani2020test}
% \href{https://www.sciencedirect.com/science/article/pii/S1361841520302711}{$^{\uparrow}$}
or in the feature space \cite{he2021autoencoder}.
% \href{https://www.sciencedirect.com/science/article/pii/S1361841521001821}{$^{\uparrow}$}.
$H_\psi$ encourages similarity between outputs or features of the test image with those of the training images. It is itself modelled using a CNN and trained in a self-supervised manner - as a denoising autoencoder (DAE) in \cite{karani2020test}
% \href{https://www.sciencedirect.com/science/article/pii/S1361841520302711}{$^{\uparrow}$}
and as an autoencoder (AE) in \cite{he2021autoencoder}.
% \href{https://www.sciencedirect.com/science/article/pii/S1361841521001821}{$^{\uparrow}$}.
% In other words, the parameters $\phi$ are adapted so that the outputs / features of the test image become similar to the outputs / features of the training images.
% The prior $H_\psi$, designed to gauge plausibility of its inputs, is itself modelled using a NN and trained in a supervised or self-supervised manner.
%Once the respective priors are trained, the following optimization is done at test-time: $min_\phi \: L(S_\theta(z), H_\psi(S_\theta(z)))$~\cite{karani2020test}\href{https://www.sciencedirect.com/science/article/pii/S1361841520302711}{$^{\uparrow}$} and $min_\phi \: L(z, H_\psi(z))$~\cite{he2021autoencoder}\href{https://www.sciencedirect.com/science/article/pii/S1361841521001821}{$^{\uparrow}$}.

% ================================
% Figure showing TTA-FoE Motivation
% ================================
\begin{figure}[t!]
\centering
    \includegraphics[trim = 0mm 110mm 0mm 00mm, angle=0, clip, width=0.75\textwidth]{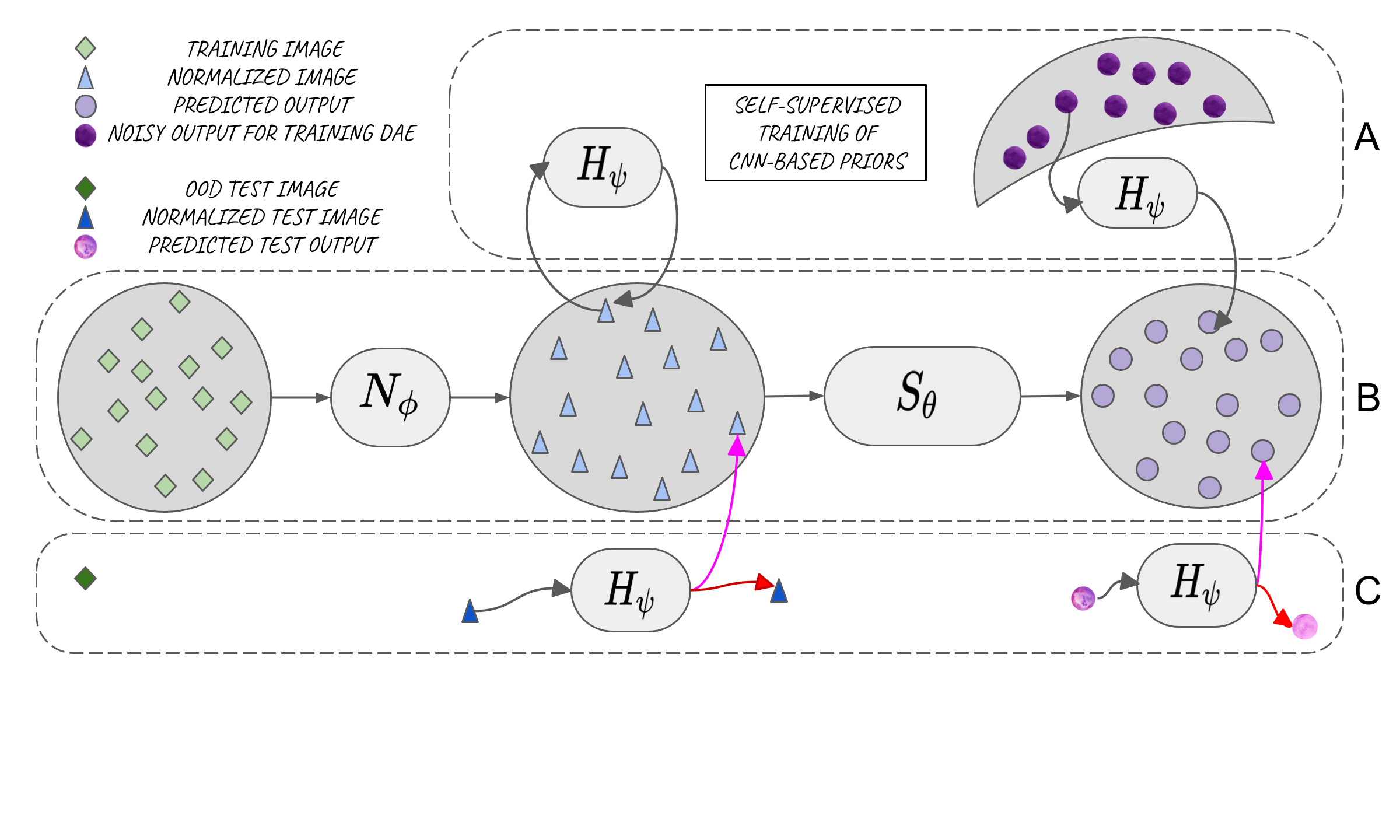}
\caption{An illustrative schematic of the DS problem in CNN-based helper models. The figure is divided into 3 horizontal slabs. Slab B shows the mapping of the inputs (green) to the outputs (purple), via the normalized features (blue). Slab A shows the training of prior models (autoencoder (center)~\cite{he2021autoencoder}, denoising autoencoder (right)~\cite{karani2020test}) to be used for TTA: the AE is trained to auto-encode features of training images and the DAE is trained to denoise corrupted outputs (from a specific corruption distribution indicated by the crescent). Finally, slab C shows the desirable behaviour (pink arrows) and potential failure cases (red arrows) when the trained prior models are used to guide TTA.}~\label{fig_tta_foe_motivation}
\vspace{-10pt}
\end{figure}

% =======================
% The DS problem in $H_{\psi}$
% =======================
\subsection{The DS problem in $H_{\psi}$}\label{subsec:ds_problem_in_tta}
% \vspace{5pt} \noindent \textbf{The DS problem in $H_{\psi}$}:
In this work, we scrutinize the prior model, $H_\psi$, which is a key component in tackling the DS problem via TTA.
Consider what happens when TTA is used to improve a CNN's prediction accuracy for an out-of-distribution (OOD) test image.
(In general, OOD images can differ from training images in terms of acquisition settings, imaging modality (e.g. CT v/s MRI), anatomy, etc. Here, we consider OOD test images pertaining to acquisition-related DS.)
At the beginning of TTA iterations, the test features (outputs) are likely to be dissimilar to the features (outputs) corresponding to the training images.
Indeed, this is symptomatic of the CNN's poor performance on OOD images.
The main assumption of TTA methods like \cite{karani2020test}, \cite{he2021autoencoder} 
is that $H_\psi$ is capable of mapping such features (outputs) to ones that are similar to features (outputs) observed during training.
However, if $H_\psi$ is modelled with a CNN, it is likely to be vulnerable to a DS problem of its own - that is, the outputs of $H_\psi$ may be unreliable when its test inputs are from a different distribution as compared to its training inputs.
An illustrative schematic of this problem is shown in Fig.~\ref{fig_tta_foe_motivation}.
AEs in~\cite{he2021autoencoder}, which are trained to auto-encode features of training images, are not guaranteed to transform the features of test images to be like the features of training images.
Similarly, DAEs in~\cite{karani2020test}, trained to denoise corrupted outputs corresponding to a particular corruption distribution, may be unable to denoise outputs with different corruption patterns.

% ================================
% Unsupervised density estimators as prior models for TTA
% ================================
\vspace{5pt} \noindent Although DAEs (for arbitrary corruption distributions) and AEs lack a strict probabilistic underpinning, the aforementioned TTA approaches can be roughly thought of as learning a probabilistic model of the training features (outputs), and then increasing the likelihood of the test features (outputs) under the trained model.
We argue that even if CNN-based unsupervised density estimation models are used as the prior, they too are likely to suffer from the DS problem \cite{nalisnick2019deep}, \cite{hendrycks2019deep}.
For instance, one approach for TTA might be to train variational autoencoders (VAEs) to model the distribution of features of the training images, and to modify the test image's features such that their likelihood under the trained VAE increases.
VAEs may even assign higher likelihood values to OOD samples than samples from their training distribution~\cite{nalisnick2019deep}.
Such behaviour may render them unsuitable for TTA.

% ================================
% Two changes proposed: (1) Match distributions instead of max. likelihood and (b) use 'SIMPLER' PRIOR MODELS
% ================================
\subsection{Overview of the proposed method}\label{subsec:method_overview}
In this work, we propose two main changes as compared to recent TTA works. %(\cite{karani2020test}\href{https://www.sciencedirect.com/science/article/pii/S1361841520302711}{$^{\uparrow}$}, \cite{he2021autoencoder}\href{https://www.sciencedirect.com/science/article/pii/S1361841521001821}{$^{\uparrow}$}).
First, instead of driving TTA by minimizing the reconstruction loss of a prior model, $H_\psi$, we propose to match the distribution of $2$D slices of a volumetric test image with the distribution of slices of training images.
The distribution matching is done in the space of normalized images, $z$.
% Indeed, distribution matching is a common UDA approach~\cite{ben2010theory}\href{https://link.springer.com/content/pdf/10.1007/s10994-009-5152-4.pdf}{$^{\uparrow}$} - although, in that setting, multiple volumetric images from the test distribution are used simultaneously.

\vspace{5pt} \noindent Second, noting the lack of DS robustness in CNN-based prior models for driving TTA, we posit that \textit{simpler} prior models may (a) suffice to improve task performance under the considered acquisition-related DS, while (b) themselves being more robust to DS as compared to CNN-based priors.
With this motivation, we model the distribution of the normalized training images, $z$, using a Field of Experts (FoEs)~\cite{roth2005fields} formulation.
FoEs (described in more detail in Sec.~\ref{sec_background}) combine ideas of Markov random fields (MRFs)~\cite{geman1984stochastic} and Product of Experts (PoEs)~\cite{hinton2002training}.
FoEs enable modeling of complex distributions as a product of several simpler distributions.
The simple distributions are those of the outputs of so-called \textit{expert} functions, which are typically formulated as scalar functions of image patches.
We propose to use the task-specific filters learned in $S_\theta$ as the FoE experts (Sec.~\ref{sec:foe_cnn_model}).
Further, we augment the FoE model with additional experts - projections onto principal components of patches in the last layer of $S_\theta$ (Sec.~\ref{sec_foe_cnn_pca_model}).

\vspace{5pt} \noindent For TTA, we adapt the normalization module $N_\phi$, so as to match the individual expert distributions of the test and training images, for all experts in the FoE model.
% Furthermore, we investigate the effect of modeling the expert distributions using parametric (Gaussian) versus non-parametric (kernel density estimation) methods.

\subsection{Summary of contributions}\label{subsec:contributions}
To summarize, we consider the acquisition-related DS problem in CNN-based medical image analyses and make the following contributions in this work:
(1) we propose distribution matching for TTA,
(2) we model the distribution of normalized images, $z$, using a FoE model, with the task-specific CNN filters acting as the expert functions, and
(3) we augment the FoE model with PCA-based expert functions.
% (4) we investigate the effect of parametric v/s non-parametric estimation of expert distributions.
% We hypothesize that the proposed FoE probability model is more robust to CNN-based helper models, $H_\psi$, and thus leads to a more stable TTA approach.

\vspace{5pt} \noindent We support these technical contributions with an extensive validation on 5 image segmentation tasks, using data from 17 centers, and an image registration task, using data from 3 centers. To the best of our knowledge, this is the first work in the literature that evaluates the TTA setting on such a large variety of anatomies and tasks for medical image analysis.
The results of these experiments help us organize the current TTA literature, including the proposed method, along three axes.
(1) Applicability to multiple tasks: some of the existing TTA methods are task-dependent. The proposed method relieves this constraint, and provides a general approach that can used in multiple tasks. As compared to existing task-agnostic methods, the proposed method provides similar performance for image registration and superior performance for image segmentation.
(2) Performance in segmentation of anomalies: we find that DS robustness issue is particularly difficult for lesion datasets. Here, all of the existing TTA methods either fail to improve performance, and several methods even lead to performance degradation as compared to the baseline. The proposed method provides substantial performance gains in this challenging scenario.
(3) Performance in segmentation of healthy tissues: in this scenario, our experiments indicate that methods specifically designed for handling distribution shifts in image segmentation outperform more general TTA methods, including the proposed method.
% First, the large-scale validation establishes the potential of TTA - overall, all compared methods show substantial robustness improvements gains for acquisition-related DS in MRI.
% Second, existing TTA methods are unstable - that is, for some test distributions, they lead to performance degradation as compared to the baseline. The proposed method does not exhibit such instability.
% =============================
% RELATED WORK
% =============================
\section{Related Work}

\subsection{Domain Generalization (DG)}~\label{sec_related_work_dg}
From a practical point-of-view, DG is arguably most attractive among all strategies for tackling the DS problem; after training, it allows a CNN to be used directly (without any adaptation) for analyzing images from unseen test distributions.
Several strategies have been proposed for DG -
(i) meta learning~\cite{dou2019domain},
(ii) domain invariant representation learning~\cite{lafarge2019learning},
(iii) shape-appearance disentanglement~\cite{liu2021semi},
(iv) regularization using task-specific priors~\cite{liu2020shape},
(v) data augmentation~\cite{zhang2019generalizing},
(vi) training with a fully-synthetic dataset of images representing a large degree of morphological, resolution and acquisition parameter variation~\cite{billot2021synthseg},
%(v) regularization to impose task-specific shape or smoothness priors~\cite{kouw2019cross}\href{https://link.springer.com/chapter/10.1007\%2F978-3-030-20351-1_63}{$^\uparrow$}~\cite{kouw2019cross}\href{https://link.springer.com/chapter/10.1007\%2F978-3-030-20351-1_27}{$^\uparrow$},
among others.
%It has also been suggested to further improve these strategies by leveraging knowledge of the image acquisition process.
%SynthSeg is a CNN trained with synthetic scans obtained by leveraging a generative model inspired from the Bayesian segmentation framework, with fully randomised parameters including aggressive augmentation and artefacts modelling.
These DG methods substantially improve CNN robustness with respect to DS; however, there still remains a gap to the performance that can be achieved if supervised learning were to be done using labelled images from the test distribution.
Further, some of these methods rely on design choices that may be applicable only for certain anatomical regions (for instance, the procedure to generate synthetic images in
\cite{billot2021synthseg} requires dense segmentation labels as inputs).
Overall, we argue that the settings of DG and TTA are complementary in nature - the former can provide a fairly robust trained model, and the latter can further improve performance by fine-tuning the model to specifically suit the test image at hand.

% =========================================================
% TEST TIME ADAPTATION
% =========================================================
\subsection{Test-Time Adaptation}
% \vspace{5pt} \noindent \textbf{Test-Time Adaptation (TTA)}:
A relatively new approach for tackling DS is to adapt a trained model using unlabelled test image(s), but without access to the training dataset.
At a broad level, works in this category vary along two axes - (a) which parameters are adapted at test time and (b) the loss function that is used to drive the adaptation.
Common choices along axis (a) are
(i) a normalization module in the task CNN's initial layers~\cite{karani2020test},~\cite{sun2020test},
(ii) batch normalization parameters throughout the task CNN~\cite{wang2020tent} and
(iii) a combination of shallow adaptable modules at different layers in the task CNN~\cite{he2021autoencoder}.
Along axis (b), proposed works either minimize
(i) the loss of a pre-trained self-supervised network~\cite{karani2020test},~\cite{he2021autoencoder},~\cite{sun2020test},
(ii) the entropy of predictions for the test image(s)~\cite{wang2020tent}, or
(iii) task-specific self-supervised losses such as
(1) k-space data consistency in MRI reconstruction CNNs~\cite{gilton2021model},
(2) cycle-consistency-based estimation of a \textit{correction filter} to transform low-resolution (LR) test images to resemble LR images seen during training of super-resolution CNNs~\cite{hussein2020correction} or
(3) an estimator (Stein's unbiased risk estimator) of the true loss for known noise distributions in denoising CNNs~\cite{soltanayev2018denoisers}.
% Two works that are the closest to the proposed method~\cite{karani2020test}\href{https://www.sciencedirect.com/science/article/pii/S1361841520302711}{$^\uparrow$}~\cite{he2021autoencoder}\href{https://www.sciencedirect.com/science/article/pii/S1361841521001821}{$^\uparrow$} have already been described in Sec.~\ref{sec:introduction}.
% Other approaches include minimizing the entropy of predictions for the test images~\cite{wang2020tent}\href{https://arxiv.org/pdf/2006.10726.pdf}{$^{\uparrow}$} and improving performance of test images on a pre-trained model for a \textit{proxy} self-supervised task such as rotation prediction~\cite{sun2020test}\href{http://proceedings.mlr.press/v119/sun20b/sun20b.pdf}{$^{\uparrow}$}.
% FROM wang2020tent: Entropy is an unsupervised objective because it only depends on predictions and not annotations. However, as a measure of the predictions it is directly related to the supervised task and model. In contrast, proxy tasks for self-supervised learning are not directly related to the supervised task. Proxy tasks derive a self-supervised label y′ from the input xt without the task label y. Examples of these proxies include rotation prediction (Gidaris et al., 2018), context prediction (Doersch et al.,2015), and cross-channel auto-encoding (Zhang et al., 2017). Too much progress on a proxy task could interfere with performance on the supervised task, and self-supervised adaptation methods have to limit or mix updates accordingly (Sun et al., 2019b;a). As such, care is needed to choose a proxy compatible with the domain and task, to design the architecture for the proxy model, and to balance optimization between the task and proxy objectives. Our entropy objective does not need such efforts.

\vspace{5pt} \noindent Test-image-specific adaptation has also been considered in the context of generative models.
For instance,~\cite{hussein2020image} proposed to fine-tune density estimation models (e.g. generative adversarial networks) for each test image, when used in the Bayesian image enhancement framework.
As well,~\cite{ulyanov2018deep} observed that CNNs trained from scratch to generate a given corrupted test image from a random vector have a tendency to first generate the corresponding clean image.
This has been recently leveraged for dynamic cardiac MRI reconstruction in~\cite{yoo2021time}.

\vspace{5pt} \noindent A closely related setting to TTA is that of source-free domain adaptation (SFDA), where multiple images from the test distribution are used simultaneously for model adaptation~\cite{jain2011online},~\cite{chidlovskii2016domain},~\cite{liang2020we},~\cite{bateson2020source}.
% As medical images from the same institution may differ in terms of their acquisition protocol parameters, we believe that TTA is better suited for medical image analysis than SFDA.
While SFDA has the advantage that multiple images from the test distribution may provide a regularization effect on one another during adaptation, TTA may benefit from adapting parameters to get the best performance for each test image. It may be interesting to empirically compare the performance of SFDA with the proposed TTA approach; we defer this analysis to future work.

% =========================================================
% BATCH NORMALIZATION STUFF
% =========================================================
\subsection{Matching Marginal Feature Distributions for Tackling DS}
% \vspace{5pt} \noindent \textbf{Adaptive Batch Normalization for Tackling DS}:
Another TTA strategy is to use the statistics of the test image(s) in the batch normalization~\cite{ioffe2015batch} layers of the task CNN. Here, no learnable parameters of the task CNN are adapted; rather the mean and variance stored in each batch normalization layer are replaced with those of the given test image(s). Effectively, at each layer, this amounts to matching the $1$D Gaussian approximation of the marginal feature distribution of the test image(s) with that of the entire training dataset.
Indeed, with this motivation,~\cite{ishii2021source} explicitly minimize the KL-divergence between Gaussian approximations of the marginal distribution of features at a particular layer in the task CNN. 
These strategies has been shown to improve DS robustness in natural imaging datasets~\cite{li2018adaptive},~\cite{schneider2020improving},~\cite{ishii2021source}.
On the other hand,~\cite{burns2021limitations} recently point out that this method matches only the first two moments of the $1$D distributions, and is thus prone to inaccuracies when the distributions are substantially non-Gaussian.
To match higher-order moments, concurrent work~\cite{eastwood2022sourcefree} matches non-parametric approximations of marginal feature distributions between test and training images.
The formulation presented in this work can be used with both parametric or non-parametric approximations.

\vspace{5pt} \noindent We believe that an important contribution of the work in this manuscript is the interpretation of marginal feature distribution matching idea in the field-of-experts (FoE) formulation.
This formulation allows us to view individual features as experts of a FoE model for the \textit{full} probability distribution of upstream features (specifically, in our case, of the normalized images, $N_\phi(X)$).
Thus, the proposed work generalizes the marginal distribution matching framework, and several previous works~\cite{li2018adaptive},~\cite{schneider2020improving},~\cite{ishii2021source},~\cite{eastwood2022sourcefree} can be seen as instances of the proposed general framework.
Furthermore, the proposed framework naturally extends to include $1$D distribution matching in the space of PCA loadings of patches of CNN features.
% While the replacement of batch normalization statistics is heuristically motivated in~\cite{li2018adaptive},~\cite{schneider2020improving}, the $1$D feature distribution matching in the proposed method emerges from a principled framework. Indeed, approaches like~\cite{li2018adaptive},~\cite{schneider2020improving} can be seen as instances of the proposed general framework. Further, the proposed framework naturally extends to include $1$D distribution matching in the space of PCA loadings.

% =========================================================
% OOD DETECTION
% =========================================================
\subsection{Frequency of summary statistics for out-of-distribution (OOD) detection}
% \vspace{5pt} \noindent \textbf{Frequency of summary statistics for out-of-distribution (OOD) detection}:
Noting that density estimation models may assign higher likelihood values to OOD samples than samples of the training distribution~\cite{nalisnick2019deep},~\cite{morningstar2021density} instead constructed $1$D PDFs of several summary statistics for the training data and evaluated the likelihood of the same statistics of test data under the constructed PDFs.
In a similar vein,~\cite{erdil2021taskagnostic} estimate $1$D marginal distributions of CNN features (using kernel density estimation) for OOD detection of MRIs. 
%The theoretical basis for the idea of feature / output space distribution matching to adapt models to shifted distributions was laid down in \cite{ben2010theory}\href{https://link.springer.com/content/pdf/10.1007/s10994-009-5152-4.pdf}{$^{\uparrow}$}. Given simultaneously access to labelled data from the training distribution as well as unlabelled images from a shifted distribution of interest, a common idea is to minimize a suitable sample-based divergence measure, $\mathcal{D}\big{[}\mathcal{P}(N_{\phi}(x)), \: \mathcal{P}(N_{\phi}(\bar{x}))\big{]}$, ($x \sim \mathcal{P}_{tr}$, $\bar{x} \sim \mathcal{P}_{ts}$), along with a task-specific loss function using input-output pairs from the training distribution \cite{ganin2016domain}\href{https://www.jmlr.org/papers/volume17/15-239/15-239.pdf}{$^{\uparrow}$}, \cite{tzeng2017adversarial}\href{https://openaccess.thecvf.com/content_cvpr_2017/papers/Tzeng_Adversarial_Discriminative_Domain_CVPR_2017_paper.pdf}{$^{\uparrow}$}.

%\vspace{5pt} \noindent In view of privacy concerns, it is prudent to develop approaches that work without need of sharing medical imaging data across institutions. In this setting, the test-time adaptation approach \cite{he2021autoencoder}\href{https://www.sciencedirect.com/science/article/pii/S1361841521001821}{$^{\uparrow}$} is to build two models: (a) a task model $y = S_\theta(N_\phi(x))$ and (b) a model of the feature distribution of training images, $H_{\psi}(z)$, where $z = N_\phi(x)$ and $x \sim \mathcal{P}_{tr}$. When a test image $\bar{x}$ from a shifted distribution is encountered, $N_{\phi}$ can be adapted to maximize the probability under $H_{\psi}$ of the $\bar{z} = N_\phi(\bar{x})$.
% =============================
% =============================
\section{Method}

% =============================
% BACKGROUND
% =============================
\subsection{Background}~\label{sec_background}
\vspace{-2ex}
% =============================
% MRFs
% =============================
\subsubsection{Markov Random Fields (MRFs)}~\label{sec_background_mrfs}
\noindent MRFs~\cite{geman1984stochastic} express a probability density function of an image, $z$, as an energy-based model:
% $p(z) = \frac{1}{\mathcal{C}} \exp ( - E(z))$,
\begin{equation}~\label{eqn_mrf1}
    p(z) = \frac{1}{\mathcal{C}} \exp ( - E(z))    
\end{equation}
where $\mathcal{C}$ is a normalization constant.
The energy of the image is defined as the sum of energies (potential functions) of all constituent $\mathcal{R}^{k\times k}$ patches (cliques),
% $z_k$: $E(z) = \sum_{k\times k\  \textrm{patches}} E(z_k)$.
$z_k$:
\begin{equation}~\label{eqn_mrf2}
    E(z) = \sum_{k \in \mathcal{K}} E(z_k)
\end{equation}
where $\mathcal{K}$ denotes the set of all $k\times k$ patches.
Typically, the energy function $E(z_k)$ is defined over relatively small patches and is hand-crafted - for instance, to encode smoothness.

% =============================
% FoEs
% =============================
\vspace{5pt} \subsubsection{Field of Experts (FoEs)}~\label{sec_background_foes}
\noindent FoEs~\cite{roth2005fields} extend the MRF idea by learning the energy function from data.
Specifically, the energy of image patches, $z_k$, is written in the Product-of-Experts (PoE) framework~\cite{hinton2002training},
\cite{welling2002learning}:
\begin{equation}~\label{eqn_poe_energy}
    E(z_k) = - \sum_{j=1}^{J} \log p(f_{j}(z_k); \alpha_j)
\end{equation}
Substituting this into Eqn.~\ref{eqn_mrf2}, the energy of the total image, z, becomes
\begin{equation}~\label{eqn_poe_energy_total}
    E(z) = - \sum_{k \in \mathcal{K}} \sum_{j=1}^{J} \log p(f_{j}(z_k); \alpha_j)
\end{equation}
The corresponding probability density function of the image, z, becomes
\begin{equation}~\label{eqn_poe_pdf}
    p(z) = \frac{1}{\mathcal{C}} \prod_{k \in \mathcal{K}} \: \prod_{j \in \textrm{experts}} p(f_{j}(z_k); \alpha_{j})
\end{equation}
Here, $f_{j}: \mathcal{R}^{k\times k} \rightarrow \mathcal{R}$ are \textit{expert} functions, and $\alpha_{j}$ are parameters of the 1D distributions of experts' scalar outputs.
The key idea in PoE and thus, FoE models is that each expert models a particular low-dimensional aspect of the high-dimensional data. Due to the product formulation, only data points that are assigned high probability by \textit{all} experts are likely under the model. In~\cite{hinton2002training},~\cite{welling2002learning},~\cite{roth2005fields}, $f_{j}$ and $\alpha_{j}$ are learned using an algorithm known as contrastive divergence, such that images in a training dataset are assigned low energy values, and all other points in the image space are assigned high energy values.

% =============================
% =============================
\vspace{5pt} \subsection{Field-of-Expert (FoE) Priors for TTA}
\begin{figure}[t!]
\centering
    \includegraphics[trim = 150mm 20mm 100mm 20mm, angle=0, clip, width=0.7\textwidth]{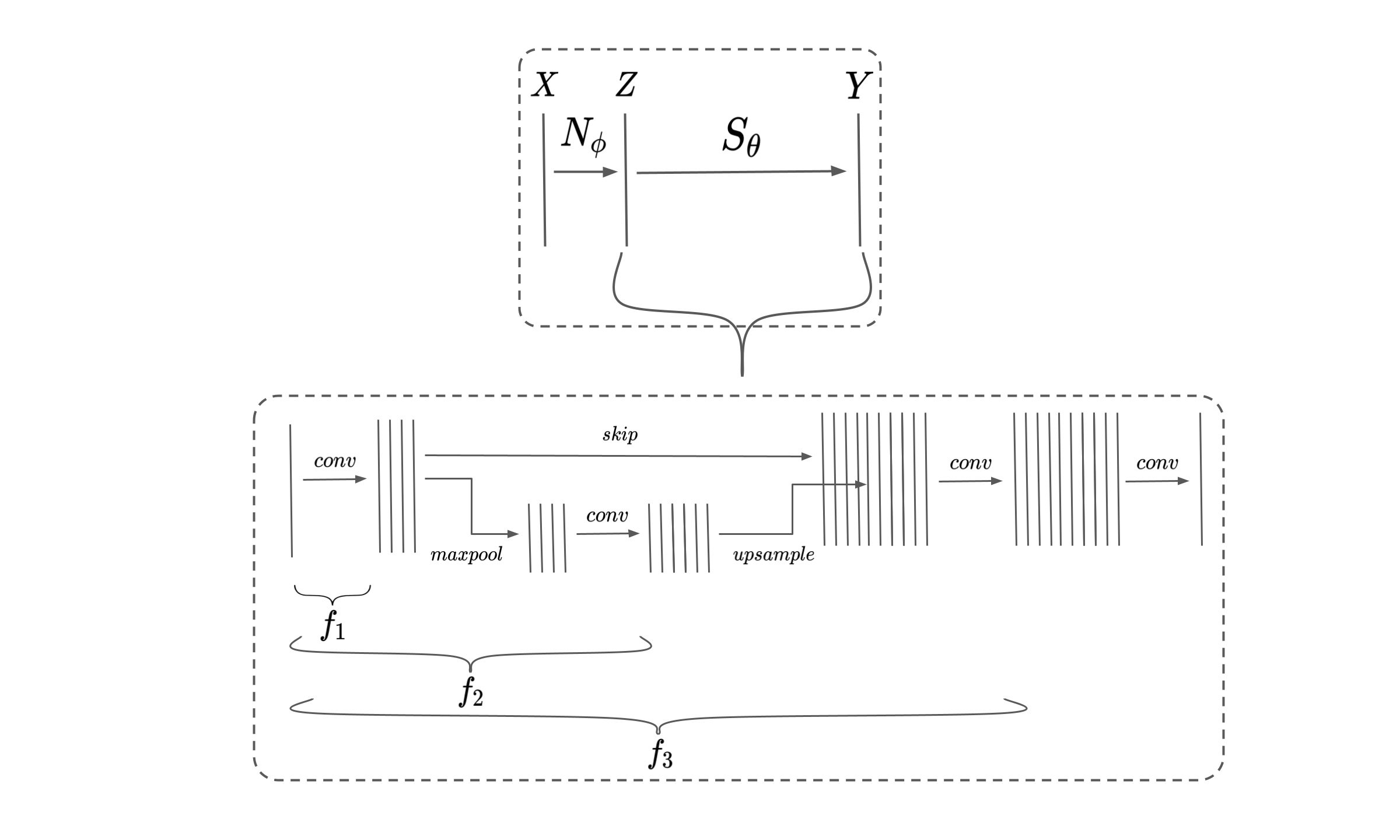}
\caption{Representative schematic of a test-time adaptable CNN.}~\label{fig_network_schematic}
\end{figure}

\noindent We now describe the proposed TTA method for acquisition-related DS in medical imaging.
Fig.~\ref{fig_network_schematic} shows a representative CNN architecture in this framework.
An image, $x$, is passed through a shallow normalization module, $N_\phi$, which outputs a normalized image, $z$.
$N_\phi$ consists of a few ($2$-$4$) convolutional layers with relatively small kernel size ($1$-$3$) and stride $1$, and outputs $z$, which is a feature with the same spatial dimensionality and the same number of channels as $x$. 
$z$ is passed through a deep CNN, $S_\theta$, which produces the output $y$. 
$y$ is formulated as per the task at hand - for instance, it can be a segmentation mask, a deformation field, a super-resolved image, etc.
We consider $2$D CNNs, but in principle, the method may be extended to $3$D architectures as well.
$S_\theta$ and $N_\phi$ are trained using labelled input-output pairs from the training distribution.
At test-time, $S_\theta$ is fixed, while $N_\phi$ is adapted for each test volumetric image.

\vspace{5pt} \noindent A representative architecture for $S_\theta$ is shown in the lower part of Fig.~\ref{fig_network_schematic}.
Let $f_l$: $\mathcal{R}^{N_x\times N_y} \rightarrow \mathcal{R}^{N_{xl}\times N_{yl}\times C_l}$ denote the function that takes as input $z$, and outputs the features of the $l^{th}$ convolutional layer of $S_\theta$.
Further, let $f_{cl}$: $\mathcal{R}^{N_x\times N_y} \rightarrow \mathcal{R}^{N_{xl}\times N_{yl}}$ denote the function that takes as input $z$, and outputs the $c^{th}$ channel of the $l^{th}$ convolutional layer of $S_\theta$.
If $k_l$ is the receptive field at $f_{cl}$ with respect to $z$, each pixel in the output of $f_{cl}$ can be seen as a $1$D projection of a $k_l \times k_l$ patch of $z$, i.e., an expert function.

% ================
% Figure of network
% ================

% ================
% DESCRIBE THE FOE MODEL OF P(Z)
% ================
\vspace{5pt} \subsubsection{The FoE-CNN model}~\label{sec:foe_cnn_model}
\noindent \noindent We model the distribution of normalized images, $z$, using the FoE formulation (Sec.~\ref{sec_background}) with the $3$ modifications.

\begin{enumerate}[label=(\roman*), wide, labelwidth=!, labelindent=0pt]

% =============================
% MULTIPLE PATCH SIZES
% =============================
\item \textbf{Multiple patch sizes}:
Firstly, note that in the original FoE model, the energy function is defined in terms of input patches of a \textit{single} patch size.
We consider multiple patch sizes to define the energy.
Specifically, if $S_\theta$ consists of $L$ convolutional layers, we consider $L$ patch sizes - namely, the receptive fields of all the convolutional layers of $S_\theta$.
\begin{equation}~\label{eqn_poe_energy_multiple_patchsizes}
    E(z) = \sum_{k=k_1}^{k_L} \: \sum_{k\times k \: \textrm{patches}} \: E((z_k))
\end{equation}

% =============================
% TASK SPECIFIC EXPERTS
% =============================
\item \vspace{5pt} \textbf{Task-specific experts}:
Secondly, we define the energy function for each patch size, using a separate PoE model.
However, unlike~\cite{roth2005fields}, we do not learn the expert functions using contrastive divergence. Instead, we construct a task-specific FoE model by using the functions $f_{cl}$ of $S_\theta$ as $C_l$ experts to describe the energy of patches of $z$ of size $k_l * k_l$:
\begin{equation}~\label{eqn_poe_energy_using_cnn_functions}
    E(z_{k_l}) = - \sum_{c=1}^{C_l} \: \log \: p(f_{cl}(z_{k_l}; \alpha_{cl}))
\end{equation}
As previously noted, $f_{cl}(z_{k_l})$ are individual pixels of the $c^{th}$ channel of the $l^{th}$ convolutional layer of $S_\theta$.
Thus, $p(f_{cl}(z_{k_l}; \alpha_{cl}))$ is the $1$D distribution of these pixel values, and $\alpha_{cl}$ are its parameters.
Combining Eqns~\ref{eqn_poe_energy_using_cnn_functions} and~\ref{eqn_poe_energy_multiple_patchsizes}, and inserting the resulting energy function into the FoE formulation (Sec.~\ref{sec_background}), the corresponding PDF of the normalized images can be written as:
\begin{equation}~\label{eqn_poe_pdf_using_cnn_functions}
    p(z) = \frac{1}{\mathcal{C}} \prod_{l=1}^{L} \: \prod_{k_l*k_l \: patches} \: \prod_{c=1}^{C_l} \\ \: p(f_{cl}(z_{k_l}))
\end{equation}

\noindent \textbf{Change of notation}: For ease of reading, let us denote expert outputs, $f_{cl}(z_{k_l})$, by $u$ and their distribution, $p(f_{cl}(z_{k_l}); \alpha_{cl})$, by $p_{cl}(u; \alpha_{cl})$.
Also, note that the product over $k_l*k_l$ patches of $Z$ is the product over the pixels of $f_{cl}$.
Thus, we have:
\begin{equation}~\label{eqn_poe_pdf_using_cnn_functions_new_notation}
p(z) = \frac{1}{\mathcal{C}} \prod_{l=1}^{L} \: \prod_{c=1}^{C_l} \: \prod_{i=1}^{N_{xl}*N_{yl}} \: p_{cl}(u_i; \alpha_{cl})
\end{equation}
The functions learned in $S_\theta$ act as \textit{task-specific} experts.
We hypothesize that matching the distributions of the outputs of such experts during TTA is likely to be beneficial for improving the task performance for the test images.
% Using distributions of different $1$D statistics has also been recently suggested for out-of-distribution detection \cite{morningstar2021density}\href{http://proceedings.mlr.press/v130/morningstar21a/morningstar21a.pdf}{$^{\uparrow}$}.

% =============================
% EXPERT DISTRIBUTIONS
% =============================
\item \vspace{5pt} \textbf{Estimation of experts' distributions}:
We approximate the expert distributions, $p_{cl}(u; \alpha_{cl})$, as $1$D Gaussian distributions, with $\alpha_{cl} = \{\mu_{cl}, \sigma_{cl}\}$:
\begin{equation}~\label{eqn_gaussian_experts}
  \begin{array}{l}
    p_{cl}(u; \alpha_{cl}) = \mathcal{N}(\mu_{cl}, \sigma_{cl}), \:
    \mu_{cl} = \frac{1}{N_z} \: \sum_{z} \: \frac{1}{N_{xl}*N_{yl}} \sum_{i} \: u_i, \:
    \sigma^2_{cl} = \frac{1}{N_z} \: \sum_{z} \: \frac{1}{N_{xl}*N_{yl}} \sum_{i} \: (u_i - \mu_{cl})^2
  \end{array}
\end{equation}
Here, the outer sum, $\sum_{z}$, is over all samples of normalized images $z$, and the inner sum, $\sum_{i}$, is over all pixels of the feature at the $c^{th}$ channel of the $l^{th}$ layer. 

\vspace{5pt} \noindent Eqn.~\ref{eqn_poe_pdf_using_cnn_functions_new_notation} defines the complete  field of CNN experts probability model (FoE-CNN) of the normalized images, $z$, with the individual expert PDFs given either by Eqn.~\ref{eqn_gaussian_experts}.

\vspace{5pt} \noindent We further analyze the effect on TTA of modelling $p_{cl}(u)$ using kernel density estimation (KDE) (see 'analysis experiments' in Sec.~\ref{sec_exp}). While this approach can capture higher-order moments of the distributions, we observed that the resulting PDFs were relatively similar to their Gaussian approximations. Thus, for simplicity, we propose to use the Gaussian approximation in the method, and show the effect of using KDE in the appendix.

\end{enumerate}

% ================
% DESCRIBE HOW TO USE THE FOE MODEL TO DRIVE TTA
% ================
\subsubsection{TTA using FoE-CNN}
\noindent We propose to use to the FoE-CNN model for TTA in the following setting: at the training site (e.g. hospital), multiple labelled volumetric images are available from the training distribution, but at the test site, we would like to adapt the model for each volumetric test image separately.
Therefore, we consider subject-specific distributions $p^s(z)$ (Eqn.~\ref{eqn_poe_pdf_using_cnn_functions_new_notation}), consisting of subject-specific $1$D PDFs, $p_{cl}^{s}(u)$ (Eqn.~\ref{eqn_gaussian_experts}).
That is, after training $N_\phi$ and $S_\theta$ using data from the training distribution, we compute and save the $1$D PDFs, $p_{cl}^{s}(u)$, for all channels of all layers, for all training subjects.
These are transferred to the test site.
A practical advantage here is that only summary statistics of the $1$D distributions are transferred - this provides benefits in terms of privacy and memory requirements, as compared to transferring large CNN models or the training distribution images themselves.
Now, for TTA, we have to make the following two design choices.

% =========================================================
% =========================================================
\begin{enumerate}[label=(\roman*), wide, labelwidth=!, labelindent=0pt]

\item \vspace{2pt} \textbf{Log-likelihood maximization v/s Distribution matching}:
Given a test subject $t$, there are two possible ways to carry out TTA.
One option is to maximize the log-likelihood of the normalized image corresponding to the test image, under the FoE-CNN model computed for the training images. Further, since the distribution of the training subjects are also modelled subject-wise, we additionally take an expectation over the training subjects:
\begin{equation}
    \begin{array}{l}
    max_\phi \: E_{p(s)} \: [E_{p^t(z)} \: \log \: p^s(z)]
    % \rightarrow max_\phi \: E_{p(s)} \: [E_{p^t(z)} \: \log \: \frac{1}{\mathcal{C}} \prod_{l=1}^{L} \: \prod_{c=1}^{C_l} \: \prod_{i=1}^{N_{xl}*N_{yl}} \: p_{cl}^s(u_i)] \\
    \rightarrow max_\phi \: E_{p(s)} \: [E_{p^t(z)} \: \sum_{l=1}^{L} \: \sum_{c=1}^{C_l} \:  \sum_{i=1}^{N_{xl}*N_{yl}} \: \log \: p_{cl}^s(u_i)]
    \end{array}
\end{equation}
We approximate the expectation with respect to $p^t(z)$ using randomly chosen $2$D slices of the test subject's volumetric image. A potential problem with this TTA formulation may be that it attract all pixels $u_i$ towards the modes of $p_{cl}^s$. To circumvent this issue, we propose to model the distribution of the normalized images corresponding to the $2$D slices of the test subject, $p^t(z)$, also using the FoE-CNN model (Eqn.~\ref{eqn_poe_pdf_using_cnn_functions_new_notation}). Now, a suitable divergence measure, $D$, between this and the distributions of the training subjects can be minimized.
\begin{equation}~\label{eqn_tta_poe_full}
    min_\phi \: E_{p(s)} \: D (p^s(z), p^t(z))
\end{equation}
However, the normalization constant $\mathcal{C}$ in Eqn.~\ref{eqn_poe_pdf_using_cnn_functions_new_notation} is intractable to compute and may be different for the two distributions. As well, commonly used divergence measures (such as f-divergences) require integration over the entire space over which the distributions are defined. Clearly, this is not possible for the high-dimensional normalized images, $z$. Therefore, for TTA, we match all the 1D expert distributions, $p_{cl}$, for all channels of all layers. That is, we minimize $L_{FoE-CNN}$ with respect to $\phi$, where
\begin{equation}~\label{eqn_tta_opt}
    \begin{array}{l}
    L_{FoE-CNN}
    = E_{p(s)} \Big{[} \: \frac{1}{L} \sum_{l=1}^{L} \frac{1}{C_l} \sum_{c=1}^{C_l} D(p_{cl}^s(u), p_{cl}^t(u)) \: \Big{]}
    \end{array}
\end{equation}
In particular, we minimize the KL-divergence between the individual $1$D distributions for the training and test images. As the $1$D PDFs are approximated as Gaussians, the KL-divergence can be computed in closed form. Further, for this choice of divergence measure, we show (Appendix Sec.~\ref{appendix_approximate_with_individual_KLs}) that minimizing the objective in Eqn.~\ref{eqn_tta_poe_full} is equal to minimizing the one in Eqn.~\ref{eqn_tta_opt} plus log of the normalization constant $\mathcal{C}$ for the test image.

% =========================================================
% =========================================================
\vspace{5pt} \item \textbf{Incorporating information from multiple training subjects}:
As mentioned previously, we consider subject-specific distributions of the normalized images.
This provides us with two options for carrying out distribution matching for TTA:
(a) minimize the \textit{divergence} of the test subject's distribution with the expected distribution over all training subjects: $min \: D(E_{p(s)}[p_{cl}^s(u)], p_{cl}^t(u))$.
(b) minimize the \textit{expected divergence} of the test subject's distribution with the distribution of each training subject: $min \: E_{p(s)}[D(p_{cl}^s(u), p_{cl}^t(u))]$.
% or $D_{KL}(E_{P(s)}[P_{cl}(u|s)], P_{cl}(u|t))$.
For KL-divergence, we show in Appendix~\ref{appendix_multiple_training_subjects} that two objectives are related as follows:
\begin{equation}~\label{eqn_kl_subjectwise_vs_average_over_subjects}
    \begin{array}{l}
        D_{KL}\big{(}E_{p(s)}[p_{cl}^s(u)], p_{cl}^t(u)\big{)} =
        - \: E_{p(s)}[D_{KL}(p_{cl}^s(u), E_{p(s)}[p_{cl}^s(u)])]
        + E_{p(s)}[D_{KL}(p_{cl}^s(u), p_{cl}^t(u))]
    \end{array}
\end{equation}
As the first term on the right-hand side of Eqn~\ref{eqn_kl_subjectwise_vs_average_over_subjects} does not depend on the test image, TTA should, in principle, be equivalent for both ways of incorporating information from multiple training subjects. However, computing (b) in practice requires only one monte-carlo (MC) approximation, while computing (a) requires three MC approximations over the training subjects. Thus, the variance of (b) will be less than that of (a)~\cite{botev2014variance}. With this reasoning, we choose (b) over (a) in the proposed TTA objective (Eqn~\ref{eqn_tta_opt}).

\end{enumerate}
\subsubsection{Additional experts using principal component analysis (PCA)}~\label{sec_foe_cnn_pca_model}
\noindent We note that the task-specific experts, $f_{cl}$, in the proposed probability model (Eqn.~\ref{eqn_poe_pdf_using_cnn_functions_new_notation}) take as inputs patches of increasing patch sizes.
The experts $f_{cL}$ have the largest receptive field, $k_L$, - thus, they model spatial correlations in $k_L\times k_L$ patches. Depending on the architecture of $S_\theta$, this may or may not cover the entire spatial dimensionality of the normalized image z. We hypothesize that considering spatial correlations in even larger image patches may further improve the proposed TTA.  Furthermore, even within the already considered patch sizes, the task-specific experts derived from $S_\theta$ may not necessarily capture all spatial correlations that are relevant for distinguishing and improving the task performance when faced with acquisition-related DS.

% https://tex.stackexchange.com/questions/129951/enumerate-tag-using-the-alphabet-instead-of-numbers
\begin{enumerate}[label=(\roman*), wide, labelwidth=!, labelindent=0pt]

% =================================
% The FoE-CNN-PCA model
% =================================
\item \textbf{The FoE-CNN-PCA model}:
We consider additional expert functions that encode spatial correlations at the layer with the largest receptive field. To do so, we use PCA~\cite{pearson1901liii},~\cite{hotelling1933analysis}.
For all the training images, we extract the last layer features, $f_{cL}(z)$.
Next, for each channel of $f_{cL}(z)$, we extract $r\times r$ patches with stride $d$.
We carry out PCA of these patches and save the first $G$ principal components.
Now, for each channel $c$, we compute the PCA coefficients, $v$, for all extracted patches of all training images.
The functions that output the PCA coefficients are considered the additional experts.
We compute subject-wise $1$D PDFs in each principal dimension, $p_{cg}^s(v)$, where $c=1,2,...C_L$, $g=1,2,...G$, $s=1,2,...n_{tr}$.

% =================================
% PCA of \textit{active} patches
% =================================
\item \vspace{2pt} \textbf{PCA of \textit{active} patches}:
For the task of image segmentation, we noticed that the marginal distributions of the features $f_{cL}$ have two distinct modes - one corresponding to the regions of interest, and one to "background" regions, which are not relevant for the task at hand.
In several segmentation applications, the background consists of many more pixels than the foreground classes combined.
In such cases, PCA may be unable to find directions of variance within the foreground regions, matching marginal distributions of which may be more useful for TTA.
To tackle this problem, we consider only \textit{active} patches while doing PCA.
Active patches are defined as those whose central pixel's predicted foreground segmentation probability is greater than a threshold $\tau$.
%The procedure remains the same as described above, except for one change - after the patch extraction step, we retain only the active patches (both for the training images as well as the test image).

% =================================
% TTA using the FoE-CNN-PCA model
% =================================
\item \vspace{2pt} \textbf{TTA using FoE-CNN-PCA}:
The principal components computed on the training images, as well as the expert PDFs of the principal coefficients are transferred to the test site.
When a test image $t$ arrives, patches of its features, $f_{cL}(z)$, are extracted, active patches are retained and the saved principal components are used to compute the corresponding expert PDFs, $p_{cg}^t(v)$.
The matching of the additional PCA coefficient PDFs is included in the TTA optimization.
That is, we minimize $L_{FoE-CNN-PCA}$ with respect to $\phi$, where
\begin{equation}~\label{eqn_tta_pca_opt}
    \begin{array}{l}
        L_{FoE-CNN-PCA} \\
        = E_{p(s)} \Big{[} \: \frac{1}{L} \sum_{l=1}^{L} \frac{1}{C_l} \sum_{c=1}^{C_l} D_{KL}(p_{cl}^s(u), p_{cl}^t(u))
        + \: \lambda \: \frac{1}{C_L} \sum_{c=1}^{C_L} \frac{1}{G} \sum_{g=1}^{G} D_{KL}(p_{cg}^s(v), p_{cg}^t(v)) \Big{]}
    \end{array}
\end{equation}
A hyperparameter, $\lambda$, is used to weigh the contribution of the PCA experts with respect to the CNN ones.

\end{enumerate}

\section{Experiments}
We validated the proposed method for tackling the DS problem on two medical image analysis tasks - segmentation (Sec.~\ref{sec_segmentation}) and atlas registration (Sec.~\ref{sec_registration}).

\subsection{Segmentation}~\label{sec_segmentation}
\vspace{-2ex}
% ================
% DATASETS
% ================
\subsubsection{Datasets}
\begin{figure}[t!]
\centering
    \includegraphics[trim = 12mm 252mm 15mm 10mm, angle=0, clip, width=0.156\textwidth]{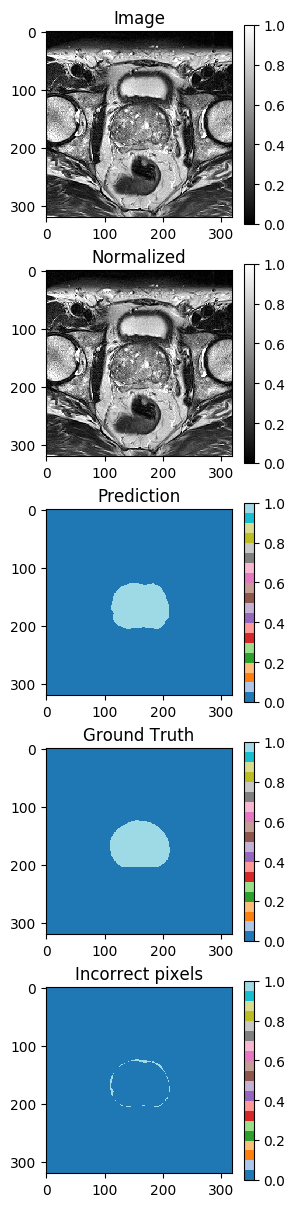}
    \includegraphics[trim = 12mm 252mm 15mm 10mm, angle=0, clip, width=0.156\textwidth]{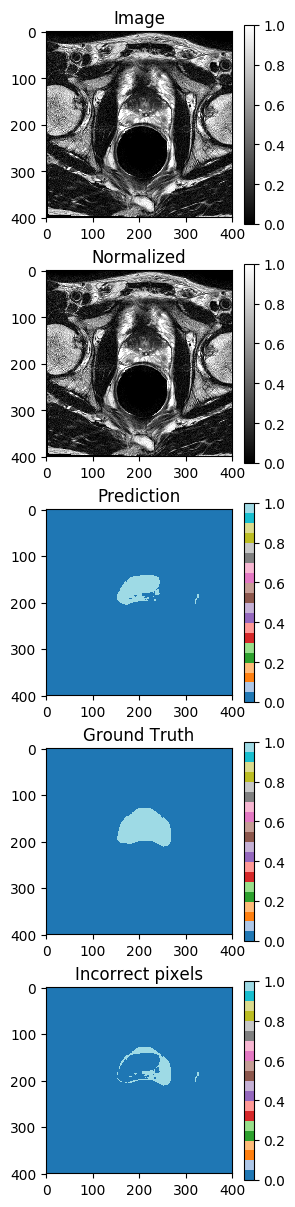}
    \includegraphics[trim = 12mm 252mm 15mm 10mm, angle=0, clip, width=0.156\textwidth]{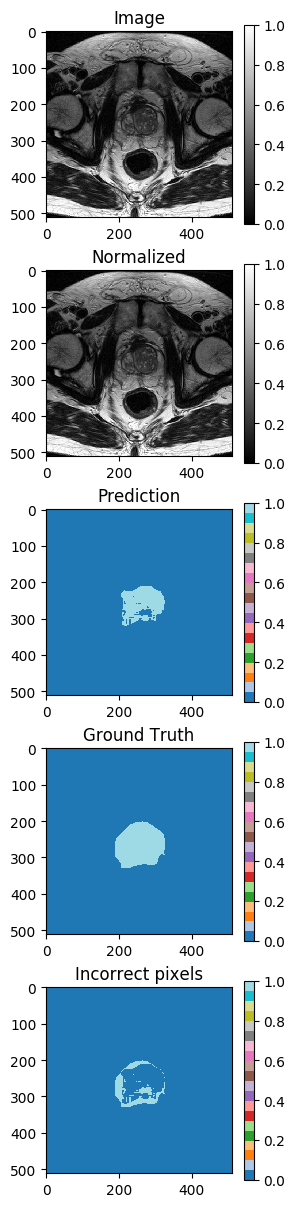}
    
    \vspace{3pt} 
    
    \includegraphics[trim = 12mm 252mm 15mm 10mm, angle=0, clip, width=0.156\textwidth]{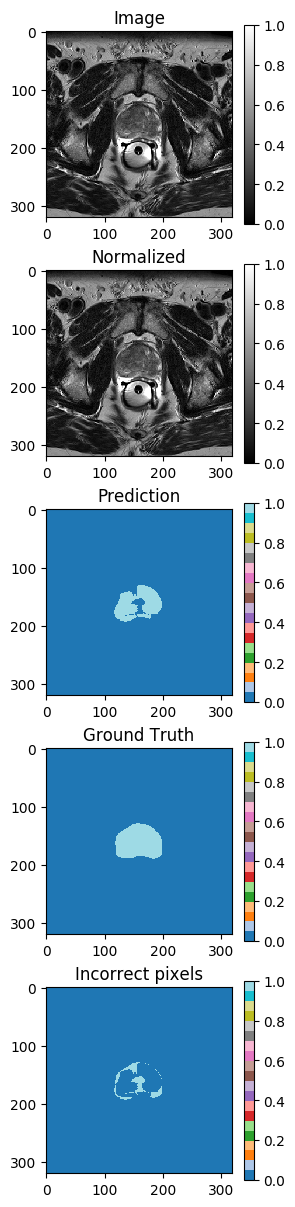}
    \includegraphics[trim = 12mm 23mm 12mm 23mm, angle=0, clip, width=0.156\textwidth]{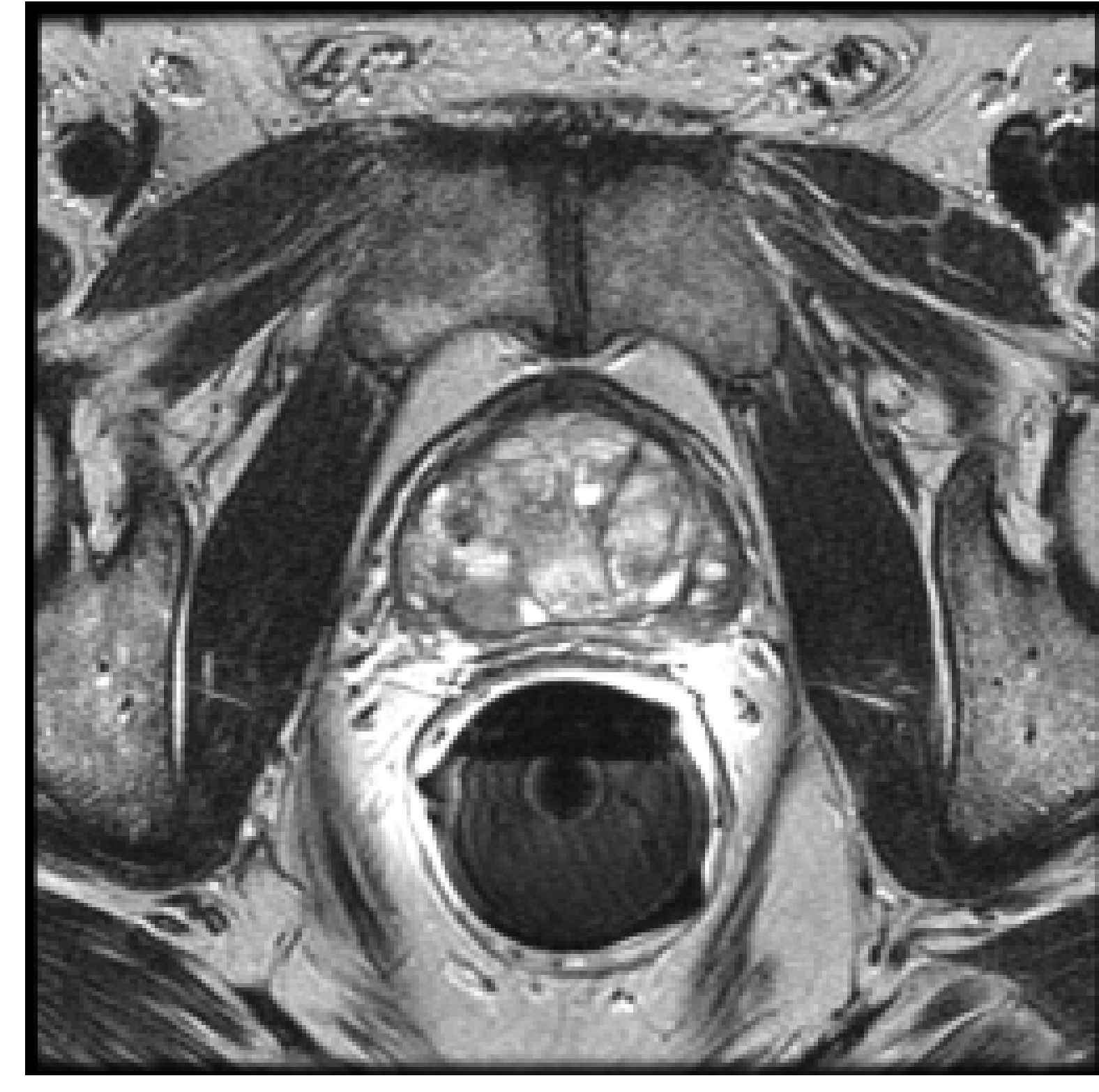}
    \includegraphics[trim = 12mm 252mm 15mm 10mm, angle=0, clip, width=0.156\textwidth]{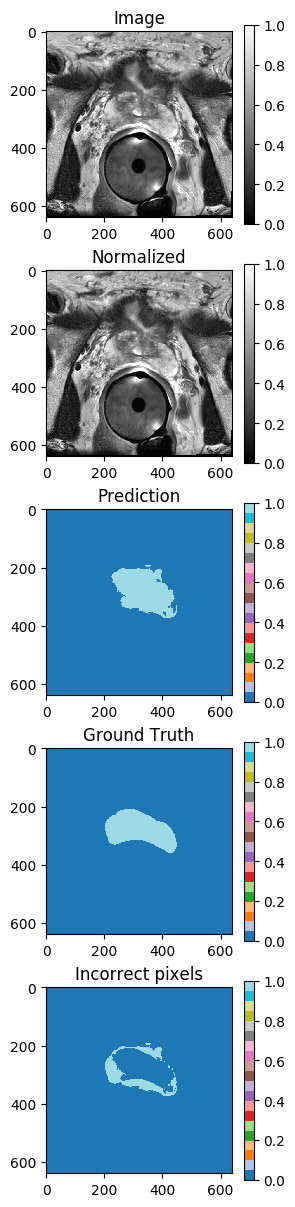}
    
    \vspace{3pt} 
    
    \includegraphics[trim = 20mm 20mm 20mm 20mm, angle=0, clip, width=0.156\textwidth]{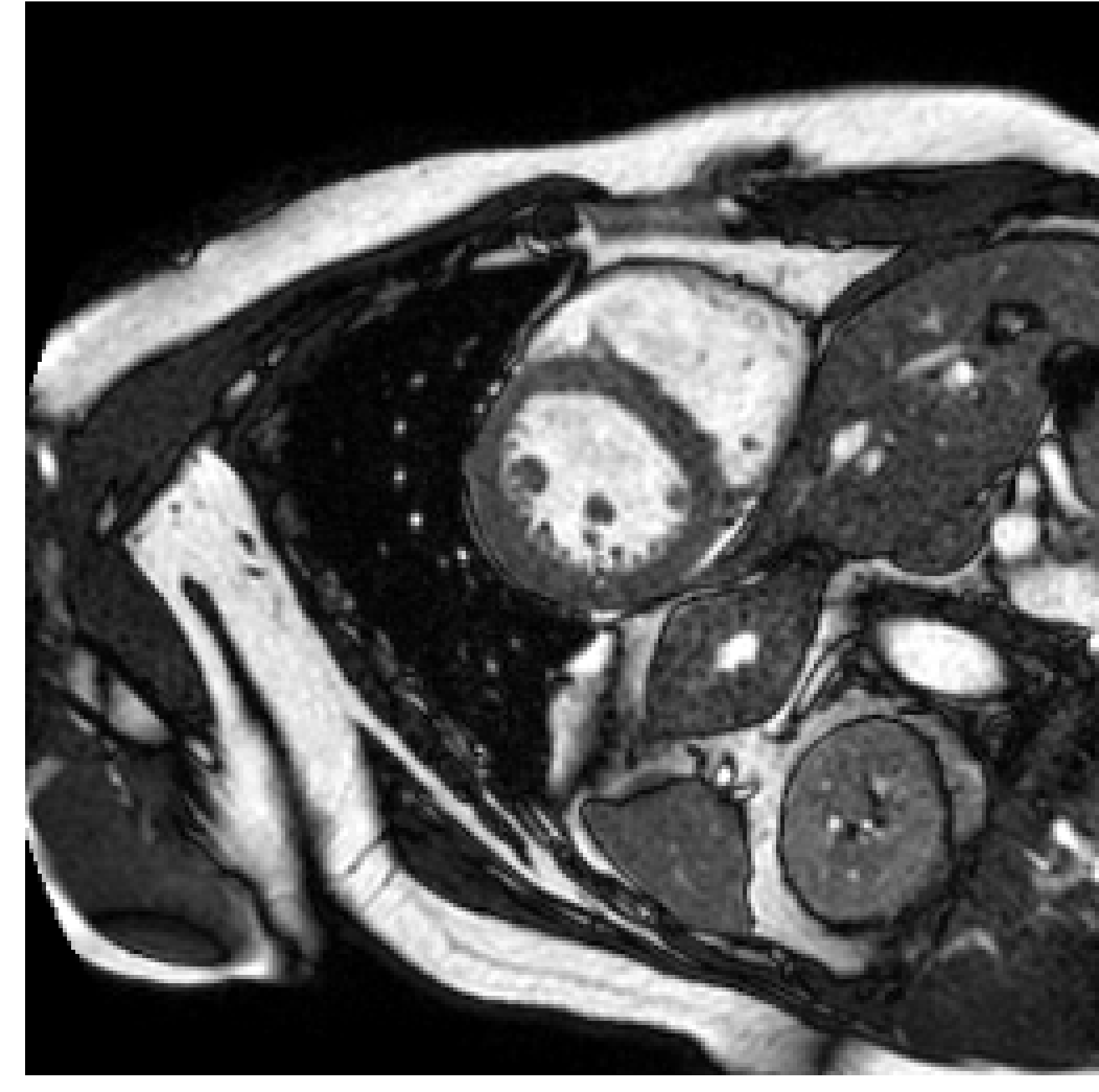}
    \includegraphics[trim = 20mm 20mm 20mm 20mm, angle=0, clip, width=0.156\textwidth]{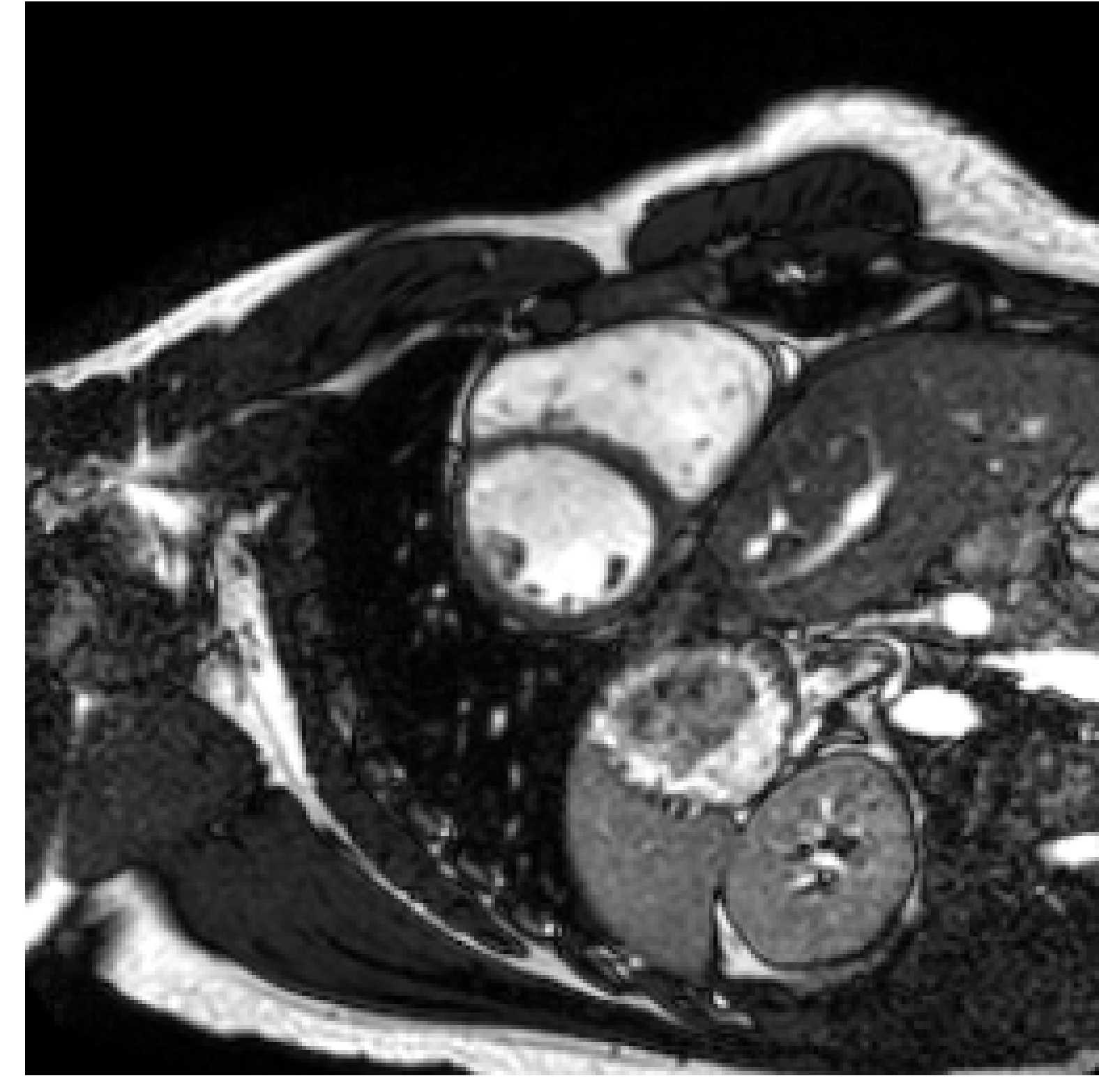}
    \includegraphics[trim = 20mm 20mm 20mm 20mm, angle=0, clip, width=0.156\textwidth]{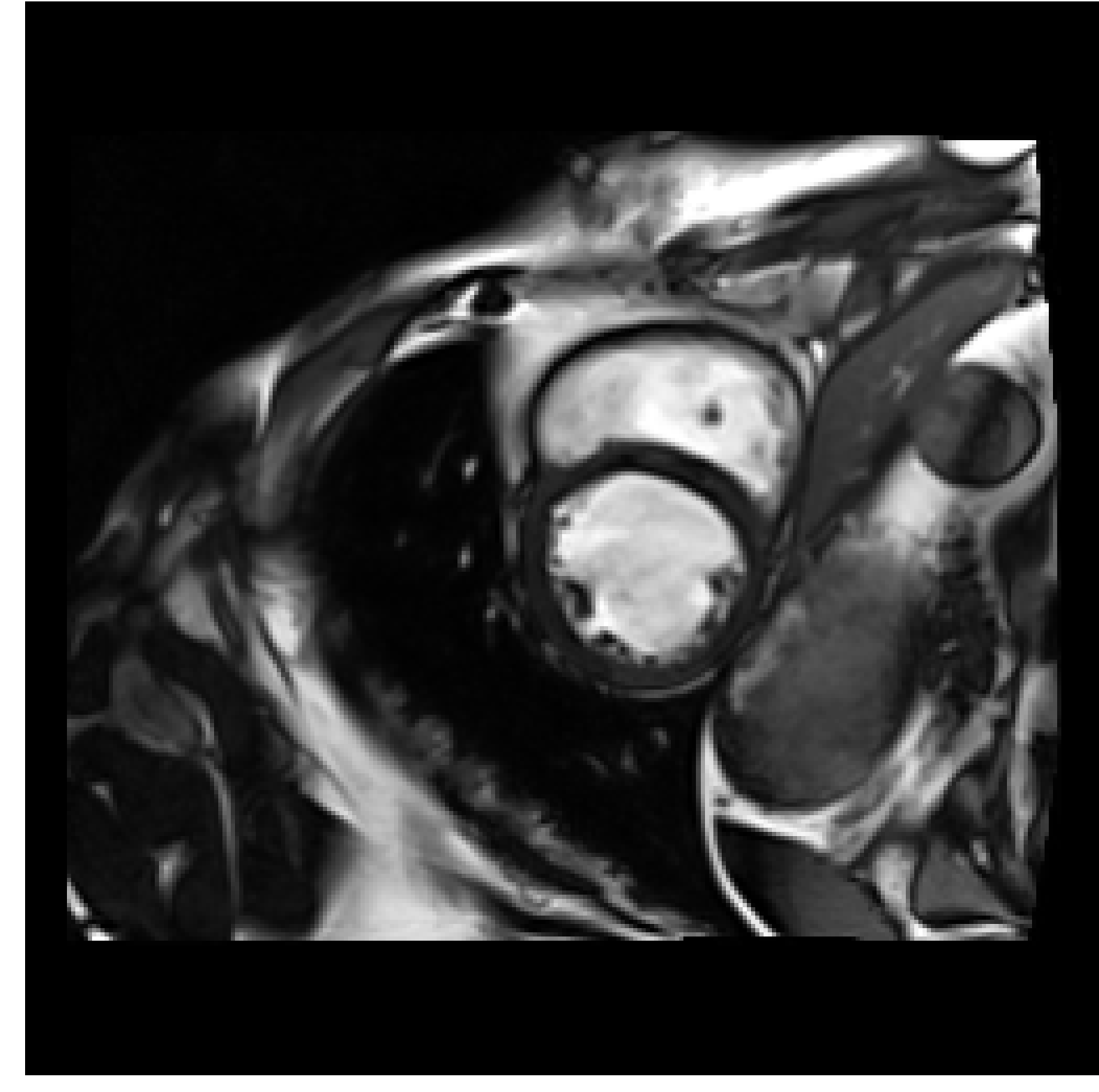}
    
    \vspace{3pt} 

    \includegraphics[trim = 5mm 5mm 5mm 5mm, angle=0, clip, width=0.117\textwidth]{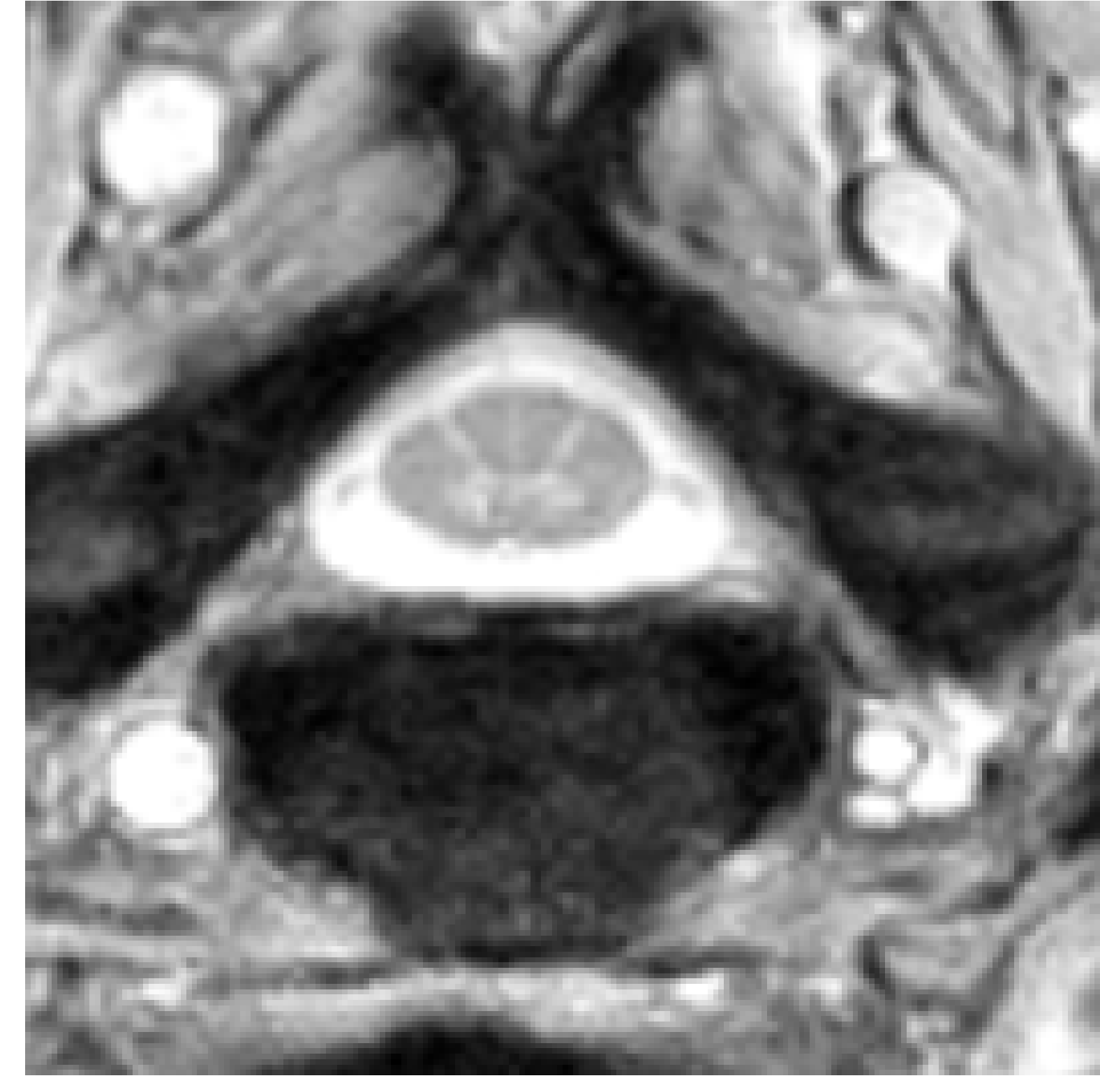}
    \includegraphics[trim = 5mm 5mm 5mm 5mm, angle=0, clip, width=0.117\textwidth]{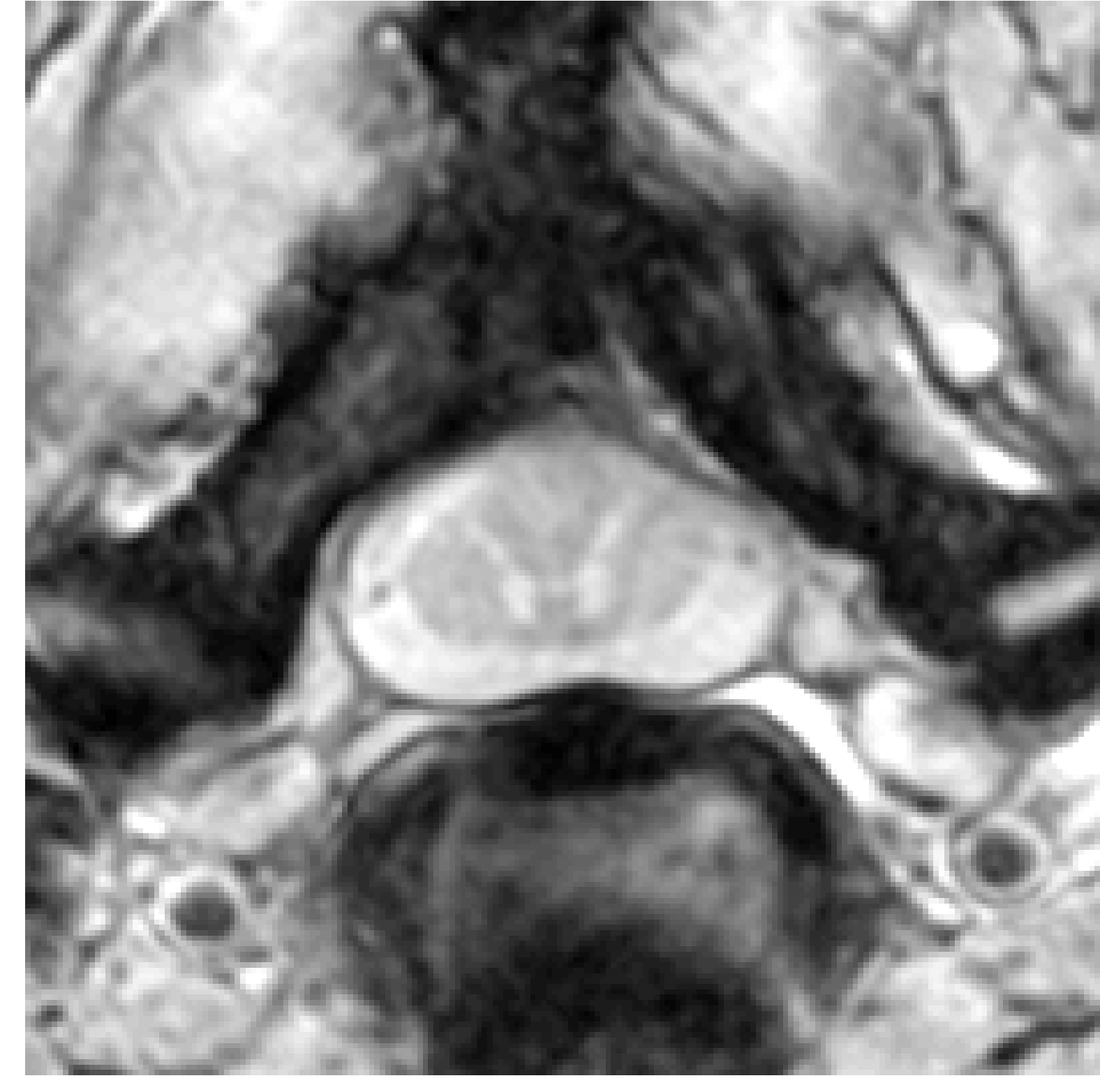}
    \includegraphics[trim = 5mm 5mm 5mm 5mm, angle=0, clip, width=0.117\textwidth]{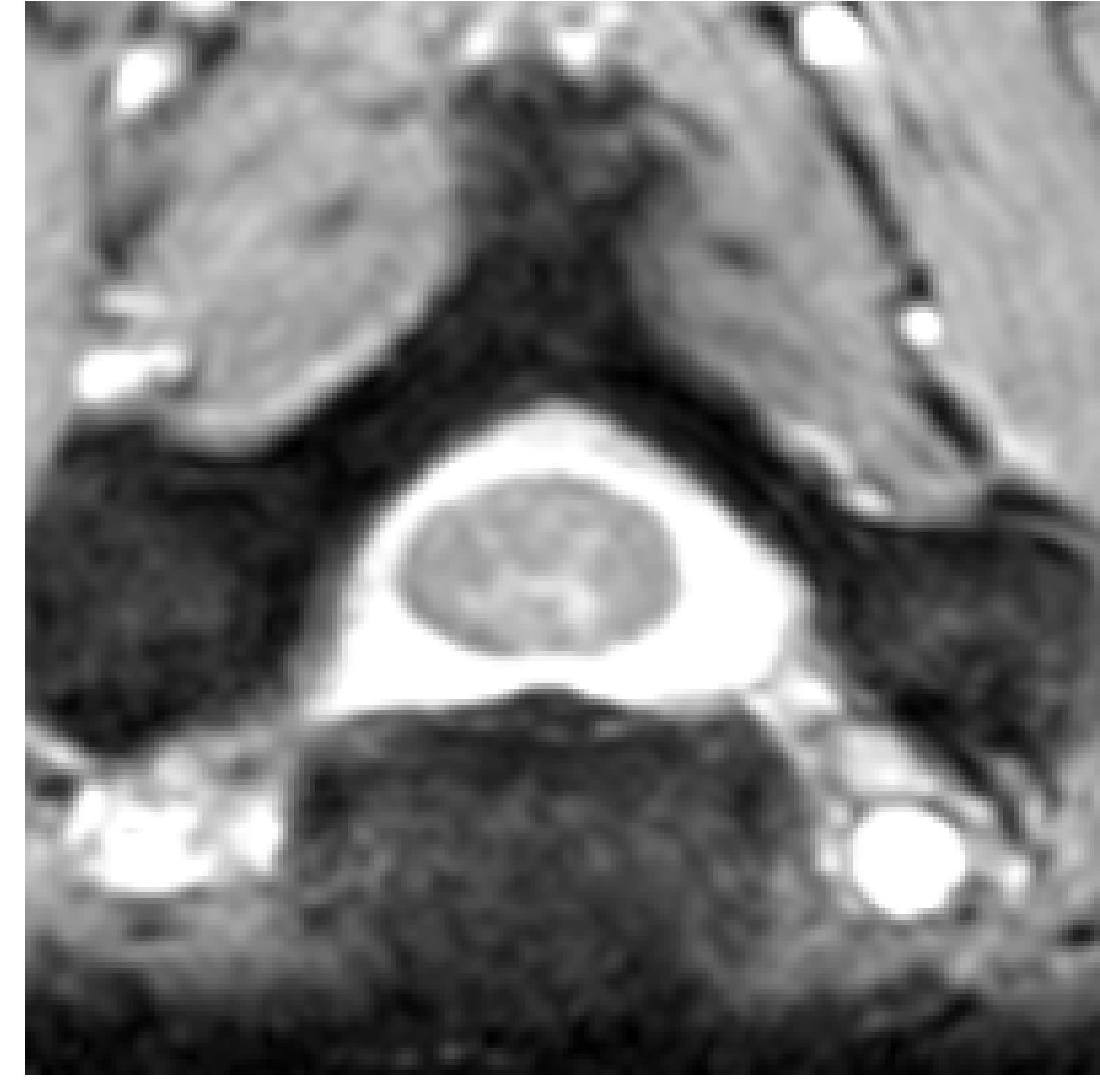}
    \includegraphics[trim = 5mm 5mm 5mm 5mm, angle=0, clip, width=0.117\textwidth]{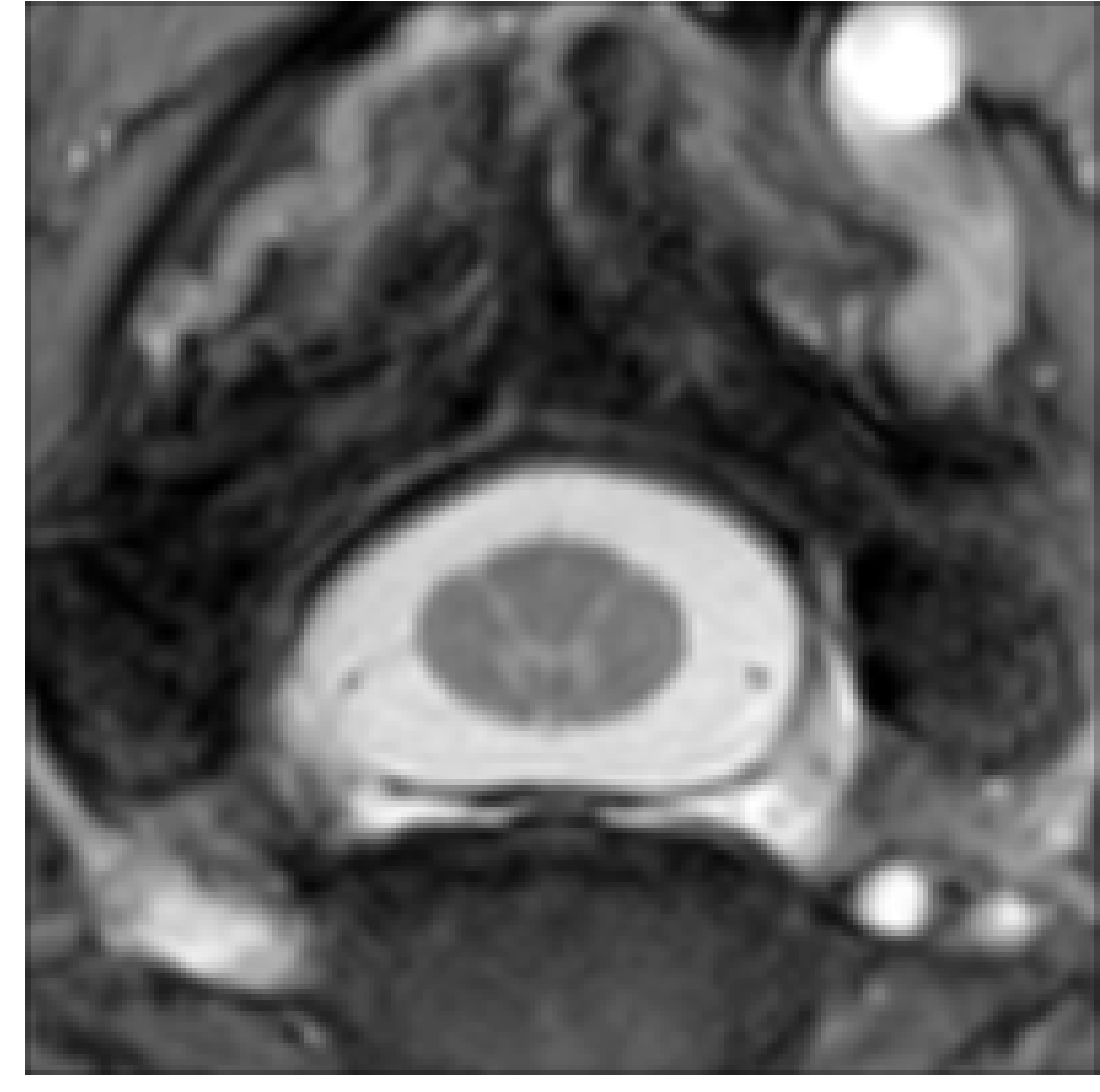}

    \vspace{3pt} 
    
    \includegraphics[trim = 15mm 5mm 15mm 25mm, angle=0, clip, width=0.117\textwidth]{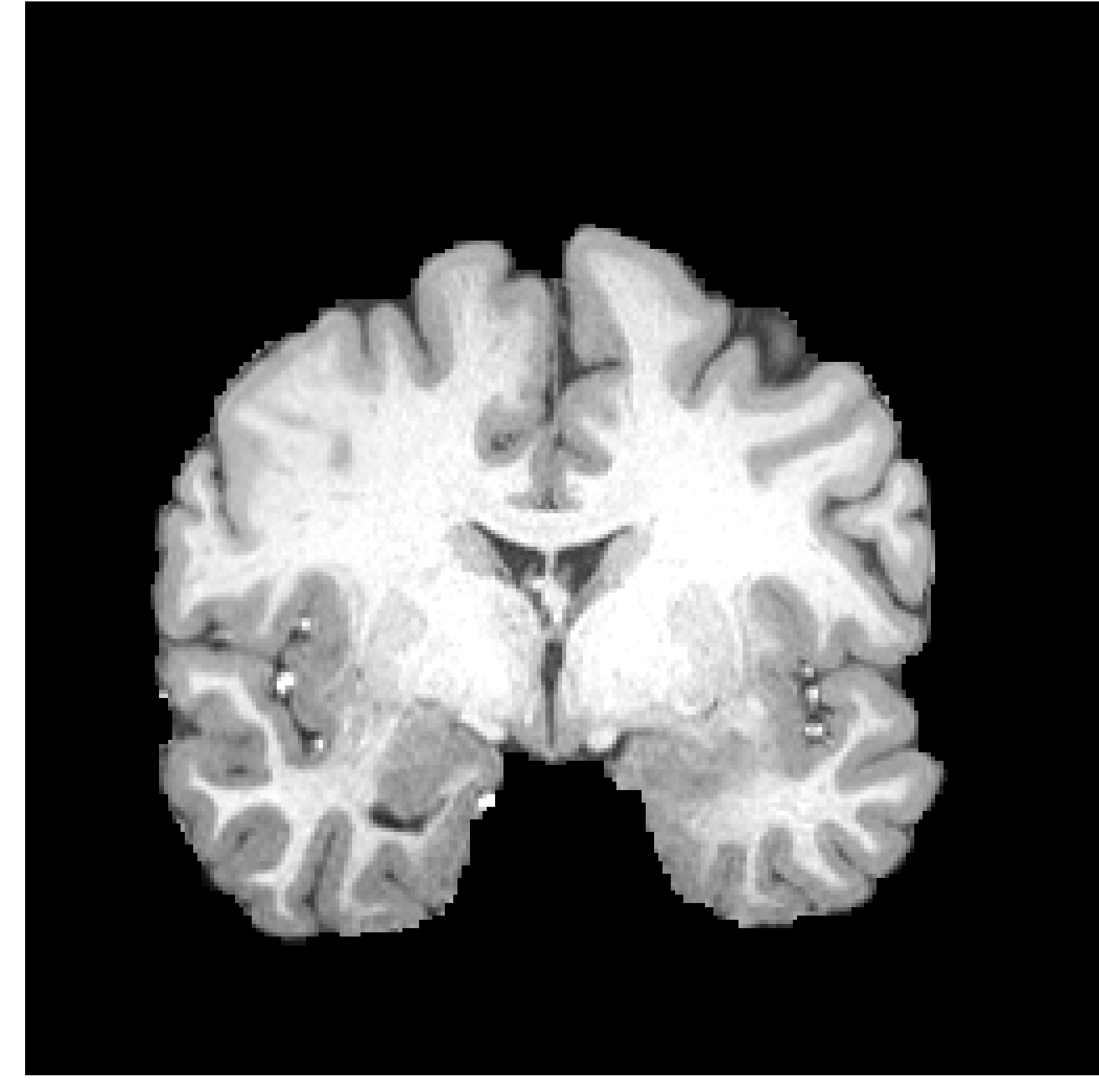}
    \includegraphics[trim = 15mm 25mm 15mm 5mm, angle=0, clip, width=0.117\textwidth]{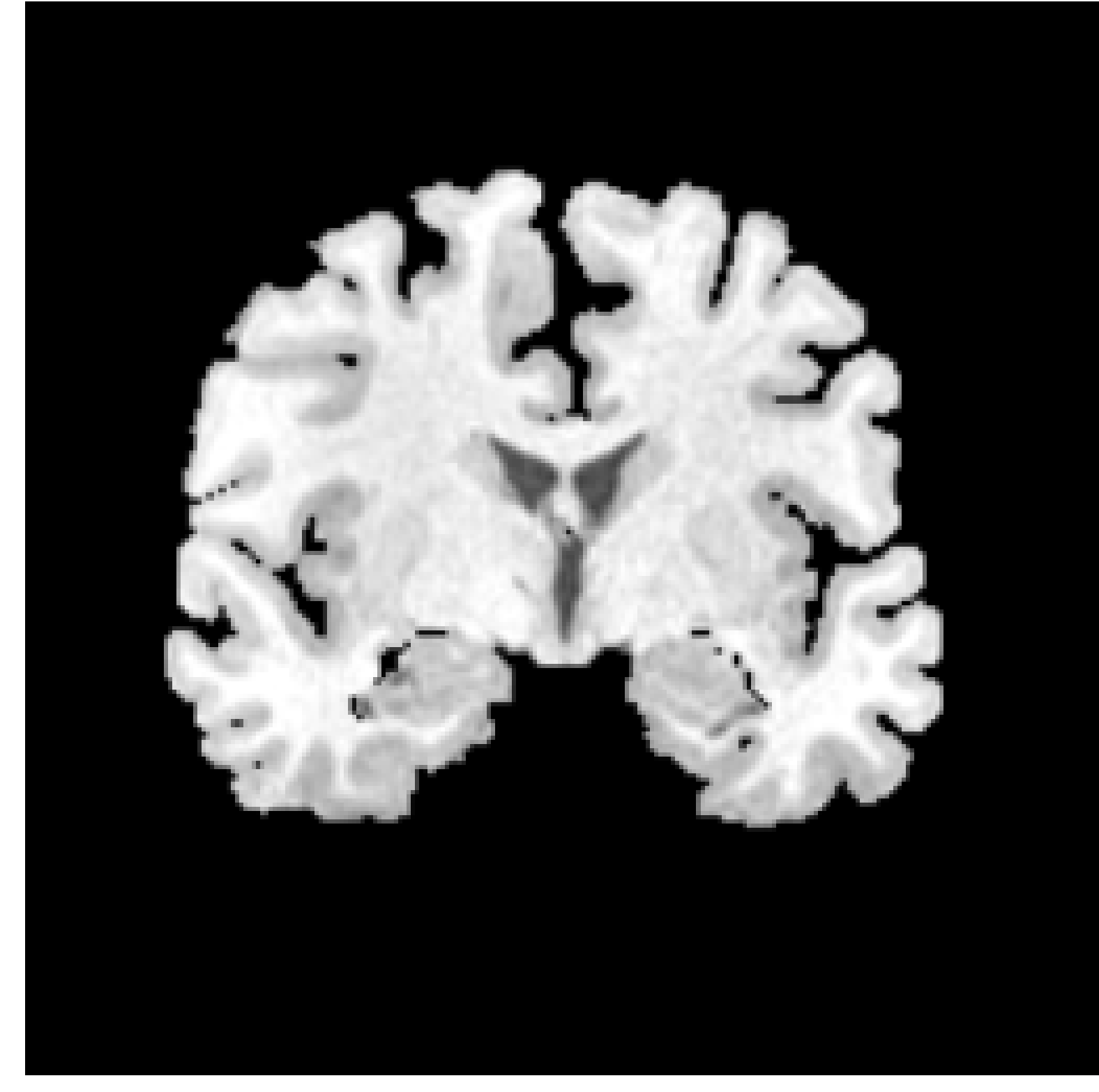}
    \includegraphics[trim = 20mm 20mm 20mm 20mm, angle=0, clip, width=0.117\textwidth]{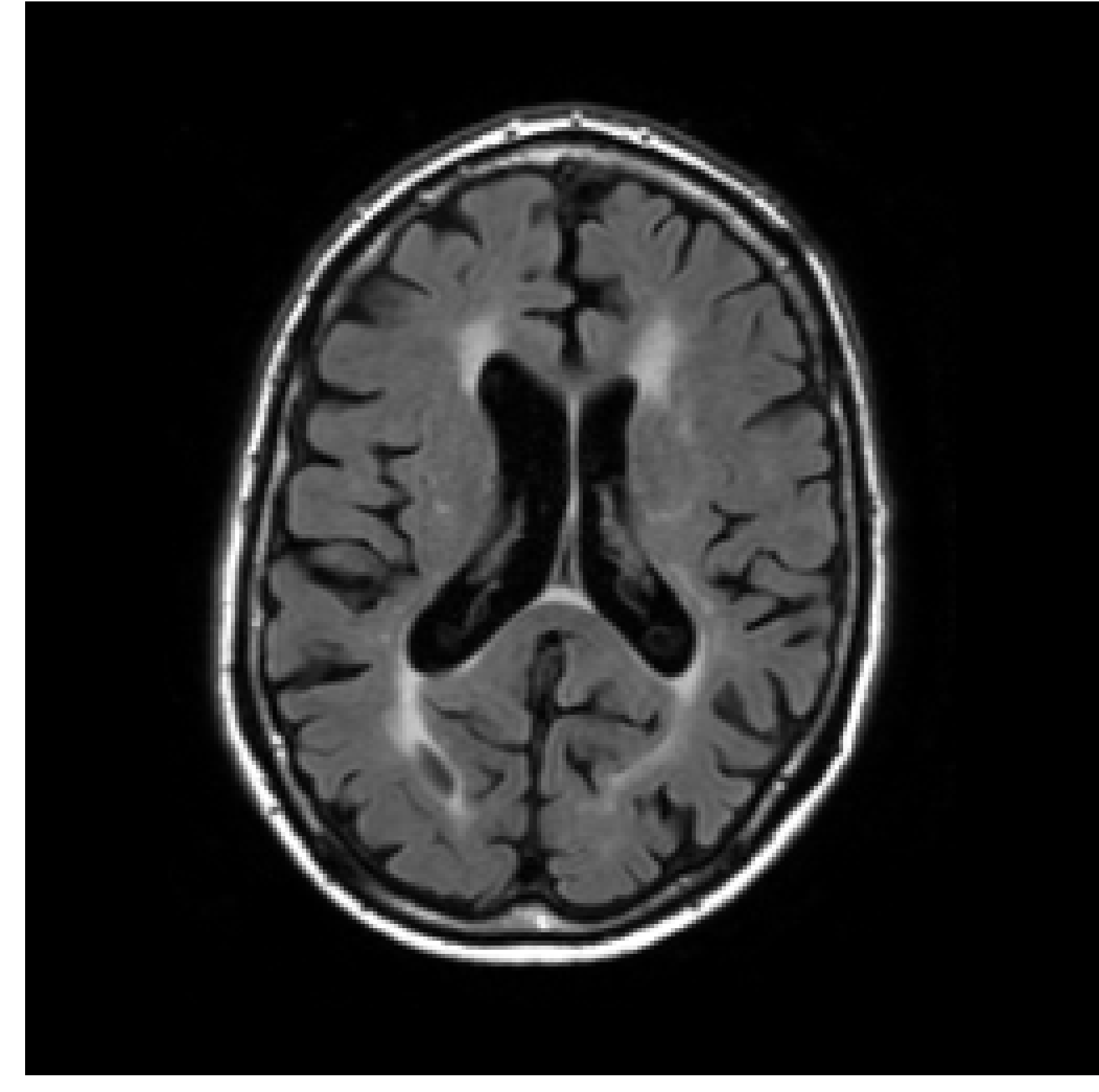}
    \includegraphics[trim = 20mm 20mm 20mm 20mm, angle=0, clip, width=0.117\textwidth]{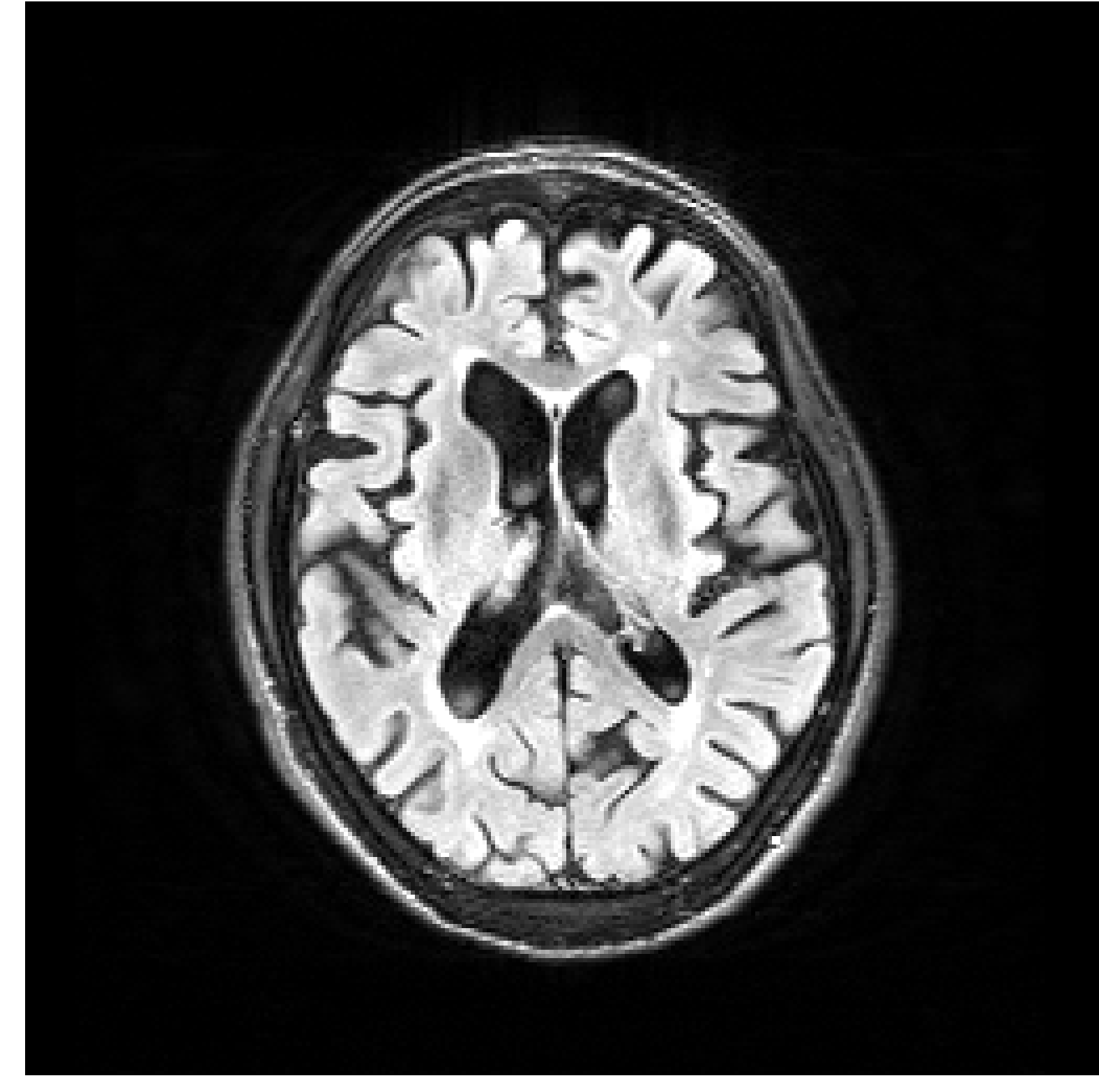}
    
\caption{Example images from the different datasets used for the segmentation experiments.
Rows 1, 2: prostate T2w MRIs from different centers (RUNMC, BMC, UCL, HK, BIDMC and USZ),
row 3: cardiac T1w images (CSF (l), UHE (c), HVHD (r)),
row 4: spine MRIs,
row 5 (first two): brain T1w MRIs of healthy subjects (HCP (l), ABIDE-CALTECH (r)), 
row 5 (last two): brain FLAIR MRIs of subjects with white matter hyperintensities (UMC (l), NUHS (r)). Please refer to Table~\ref{tab:dataset_details_seg} for details about the imaging protocol differences.}~\label{fig_dataset_details_seg}
\end{figure}

\noindent We considered MRI segmentation for 5 anatomies (names of the segmented foreground classes are shown brackets) -
(i) T2w prostate (whole organ),
(ii) Cine cardiac (myocardium, left and right ventricles),
(iii) T1w spine (spinal cord grey matter),
(iv) healthy T1w brain (cerebellum gray matter, cerebellum white matter, cerebral gray matter, cerebral white matter, thalamus, hippocampus, amygdala, ventricles, caudate, putamen, pallidum, ventral DC, CSF and brain stem) and
(v) diseased FLAIR brain (cerebral white matter hyper-intensities).
In total, we used data from 17 centers.
We used FreeSurfer~\cite{fischl2012freesurfer} generated ground truth segmentations for the healthy brain images, while expert manual ground truth annotations were available for all other datasets.
Table~\ref{tab:dataset_details_seg} shows details of all datasets; Fig.~\ref{fig_dataset_details_seg} shows example images.

\begin{table}[b!]
    \scriptsize
    \centering
    \begin{tabular}{|c|c|c|c|c|c|}
        
        \hline
        \rowcolor{titlegray} 
        Dataset & Center & Vendor & Field & $N_{I}$ & $N_{tr}|N_{vl}|N_{ts}$\\
        
        \hline
        \rowcolor{lightgray} \multicolumn{6}{|c|}{{{Prostate
        \cite{bloch2015nci},
        \cite{litjens2014evaluation},
        \cite{becker2017direct}}}} \\
        \hline
        \rowcolor{lightergray}
        NCI-13 & RUNMC, Nijmegen & S & $3$ & $30$ & $15|5|10$ \\
        \rowcolor{lightergray}
        NCI-13 & BMC, Boston & P & $1.5$ & $30$ & $15|5|10$ \\
        \rowcolor{lightergray}
        Promise12 & UCL, London & S & $1.5$, $3$ & $13$ & $(6|2|5)$x$2$ \\
        \rowcolor{lightergray}
        Promise12 & HK, Bergen & S & $1.5$ & $12$ & $(5|2|5)$x$2$ \\
        \rowcolor{lightergray}
        Promise12 & BIDMC, Boston & G & $3$ & $12$ & $(5|2|5)$x$2$ \\
        \rowcolor{lightergray}
        Private & USZ, Zurich & S & $3$ & $68$ & $48|10|10$ \\
        
        \hline
        \rowcolor{lightgray} \multicolumn{6}{|c|}{{{Cardiac
        \cite{campello2021multi}}}} \\
        \hline
        \rowcolor{lightergray}
        M\&Ms & CSF, Barcelona & P & $1.5$ & $50$ & $30|10|10$ \\
        \rowcolor{lightergray}
        M\&Ms & UHE, Hamburg & P & $1.5$ & $25$ & $10|5|10$ \\
        \rowcolor{lightergray}
        M\&Ms & HVDH, Barcelona & S & $1.5$ & $75$ & $55|10|10$ \\
        
        \hline
        \rowcolor{lightgray} \multicolumn{6}{|c|}{{{Spinal Cord Grey Matter
        \cite{prados2017spinal}}}} \\
        \hline
        \rowcolor{lightergray} SCGM & PM, Montreal & S & $3$ & $10$ & $(5|2|3)$x$3$ \\ 
        \rowcolor{lightergray} SCGM & USZ, Zurich & S & $3$ & $10$ & $(5|2|3)$x$3$ \\ 
        \rowcolor{lightergray} SCGM & VU, Nashville & P & $3$ & $10$ & $(5|2|3)$x$3$ \\
        \rowcolor{lightergray} SCGM & UCL, London & P & $3$ & $10$ & $(5|2|3)$x$3$ \\
        
        \hline
        \rowcolor{lightgray} \multicolumn{6}{|c|}{
        Brain~\cite{van2013wu},~\cite{di2014autism}} \\
        \hline
        \rowcolor{lightergray} HCP & HCP, Missouri & S & 3 & $35$ & $20|5|10$ \\
        \rowcolor{lightergray} ABIDE & AC, Caltech & S & 3 & $25$ & $10|5|10$ \\
        % \rowcolor{lightergray} ABIDE & AS, Stanford & X & x & $25$ & $10|5|10$ \\
        
        \hline        
        \rowcolor{lightgray} \multicolumn{6}{|c|}{{{White Matter Hyperintensities~\cite{kuijf2019standardized}}}} \\
        \hline
        \rowcolor{lightergray} WMH-17 & UMC, Utrecht & P & $3$ & $20$ & $(10|5|5)$x$2$ \\
        \rowcolor{lightergray} WMH-17 & NUHS, Singapore & S & $3$ & $20$ & $(10|5|5)$x$2$ \\
        
        \hline
        
    \end{tabular}
    \caption{Details of segmentation datasets for 5 anatomies.
    The vendors S, P and G refer to Siemens, Philips and GE, respectively.
    $N_I$ refers to the total number of 3D images, and the last column refers to the training, validation and test split.
    For some datasets, the split is followed by x$2$ or x$3$.
    This refers to the number of dataset splits that were done to get a reasonable number of test images in datasets with a low $N_I$.
    Among the prostate datasets, RUNMC and UCL were acquired with surface coil, while the rest of the datasets were acquired with endo-rectal coil.}~\label{tab:dataset_details_seg}
    \vspace{-10pt}
\end{table}

% ================
% PRE-PROCESSING
% ================
\subsubsection{Pre-processing}
\noindent We pre-processed all images by
(a) removing bias fields with the N4 algorithm~\cite{tustison2010n4itk},
(b) linearly normalizing the intensities to $0$-$1$ range using the $1^{st}$ and $99^{th}$ percentile values, and clipping the values to $0$ and $1$,
(c) re-scaling all images of a particular anatomy to the same in-plane isotropic resolution: $0.625 \: mm^2$, $1.33 \: mm^2$, $0.25 \: mm^2$, $0.7 \: mm^2$ and $1.0 \: mm^2$ for prostate, cardiac, spine, brain and WMH respectively,
and (d) cropping / padding zeros to have the same in-plane image size: $200$x$200$ for the spine images and $256$x$256$ for other anatomies.
The evaluation for each test image was done in its original resolution and size.
For the brain datasets, we additionally set the intensities of the skull voxels to 0.

% ================
% EXPERIMENTS
% ================
\vspace{5pt} \subsubsection{Experiments}~\label{sec_exp}

\noindent We used the same architecture for $N_\phi$ and $S_\theta$ as in~\cite{karani2020test}. $N_\phi$ consisted of $3$ convolutional layers of kernel size $3$, number of output channels $16$, $16$ and $1$, and an expressive activation function ($act(x) = \exp-(x^2/\sigma^2)$) with a learnable scale $\sigma$ for each channel. $S_\theta$ followed a U-Net~\cite{ronneberger2015u} like encoder-decoder structure with skip connections, and batch normalization layers following each convolutional layer. The ReLU activation function was used in $S_\theta$.

\vspace{5pt} \noindent For each anatomy, we used the institution in the first row in Table~\ref{tab:dataset_details_seg} as the training distribution, and the remaining institutions as separate test distributions. In this setup, we carried out the following experiments:

\begin{enumerate}[label=(\roman*), wide, labelwidth=!, labelindent=0pt]
    
    \item \textbf{Baseline}: We trained a CNN ($N_\phi$ + $S_\theta$) using labelled images from the training distribution. The training was done by minimizing the Dice loss~\cite{milletari2016v} using an Adam optimizer with a learning rate of $0.001$ and a batch size of $16$. The optimization was run for $30000$ iterations, and the model selection criterion was the average Dice score on the validation dataset. For datasets where the total number of images was very small, splits were created as indicated in Table~\ref{tab:dataset_details_seg}, and average test scores are reported. The dataset splits were designed in such a way that we had $10$ test volumes from each test distribution (except for the spine images, where the number of test volumes was $9$).
    
    \item \textbf{Strong baseline}: As described in Sec.~\ref{sec_related_work_dg}, several domain generalization methods have been proposed to tackle acquisition-related DS in medical image analysis. We implemented stacked data augmentations~\cite{zhang2019generalizing}, which present an effective and general DG approach. The implementation details were the same as in~\cite{karani2020test}: for every image in a training batch, each transformation (translation, rotation, scaling, elastic deformations, gamma contrast modification, additive brightness and additive Gaussian noise) was applied with probability $0.25$. This functioned as a \textit{strong baseline}, the performance of which we sought to improve with the proposed TTA approach.
    
    \item \textbf{Benchmark}: The best performance on images from a test distribution can be achieved by training a new model in a supervised manner, using a separate set of labelled images from the test distribution. As some of the datasets contained only a small number of images to start with, we instead used a \textit{transfer learning} benchmark - that is, the model trained on the training distribution was fine-tuned using labelled images from the test distribution. The fine-tuning was done with the Adam optimizer for $5000$ iterations, with a learning rate of $0.0001$ and batch size of $16$. This model served as the benchmark.
    
    \item \textbf{Test-Time Adaptation}: We compared the proposed approach (TTA-FoE-CNN-PCA) with four existing TTA works: TTA-Entropy-Min~\cite{wang2020tent}, TTA-DAE~\cite{karani2020test}, TTA-AE~\cite{he2021autoencoder} and TTA-FoE-CNN~\cite{eastwood2022sourcefree}.
    
    \vspace{2pt} \noindent Using the \textit{strong baseline} model as the starting point, TTA was run for $N_{tta}$ epochs for each test subject. In each epoch, averaged gradients over batches of size $b_{tta}$ were used to update the network parameters with a learning rate of $lr_{tta}$. $N_{tta}$ was set to $200$ for the healthy brain dataset (due to its high through-plane size) and to $1000$ for all other datasets. $b_{tta}$ was set to $8$ for all datasets except SCGM, where it was set to $2$ as some images had less than $8$ slices. Specific implementation details for each TTA method are provided below.
    
    \begin{enumerate}[wide, labelwidth=!, labelindent=0pt]
        \item \textbf{TTA-Entropy-Min}~\cite{wang2020tent}: The normalization module, $N_\phi$, was adapted for each test subject, with $lr_{tta}$ as $0.0001$.
        
        \item \textbf{TTA-DAE}~\cite{karani2020test}: A $3$D denoising autoencoder was trained in the space of segmentation labels, using the same corruption distribution as proposed in the original paper. Similar to the original implementation, healthy brain segmentations were downsampled in the through-plane direction by a factor of $4$, to overcome memory issues. $lr_{tta}$ was set to $0.001$.
        
        \item \textbf{TTA-AE}~\cite{he2021autoencoder}: Instead of adapting $N_\phi$, \textit{adaptor} modules $A^x$, $A^1$, $A^2$ and $A^3$ were introduced and adapted for each test subject as was done in the original article. We experimented with different settings of~\cite{he2021autoencoder} so as to get the best results for the datasets used in our experiments (Appendix~\ref{appendix_tta_ae_variants}). The architectures of the adaptors were kept the same as proposed in~\cite{he2021autoencoder}, with one change: the instance normalization layers in $A_X$ were discarded as they lead to instability during TTA. Two other changes were done to further improve the performance and stability: (a) average gradients over all batches in a single TTA epoch were used for the TTA updates (as described in Sec 3.5 in~\cite{karani2020test}) and (b) the $lr_{tta}$ was set to $0.00001$. Five $2$D autoencoders (AEs) (with the same architectures as in~\cite{he2021autoencoder}) were trained and the weight of the orthogonality loss, $\lambda_{orth}$, was set to $1.0$, as done in~\cite{he2021autoencoder}. We observed that driving the TTA using losses from two AEs (at the input and output layers) provided better performance than using all 5 AEs. With these modifications, TTA-AE worked in a stable manner, without having to resort to early stopping as done in~\cite{he2021autoencoder}.
        
        \item \textbf{TTA-FoE-CNN}: At the end of the \textit{strong baseline} training, the FoE-CNN model was constructed by computing $1$D PDFs for all channels of all layers of $S_\theta$, for each training subject. For the chosen architecture of $S_\theta$, this amounted to $704$ channels. As the PDFs were approximated as Gaussians, two parameters were stored per PDF. In principle, this method resembles the approach proposed in~\cite{eastwood2022sourcefree}.
        
        \item \textbf{TTA-FoE-CNN-PCA} For computing the additional expert PDFs of the FoE-CNN-PCA model, the following steps were followed:
        (a) For all training images, features from the last layer of $S_\theta$ were extracted (from here, a $1$x$1$ convolutional layer provided the segmentation logits). In the chosen architecture, these features were of the same spatial dimensions as the images and had $C_L = 16$ channels.
        (b) For each channel in these features, patches of size $r\times r = 16\times 16$ were extracted with stride $d=8$.
        (c) From these, only \textit{active} patches (that is, patches whose central pixel's predicted foreground probability was greater than $\tau = 0.8$) were retained. As CNNs typically make high confidence predictions, this step is likely to be insensitive to the exact value of $\tau$. To obtain a comparable number of \textit{active} patches to other anatomies, the stride $d$ was set to $2$ for the WMH images, where the foreground size was particularly small.
        (d) PCA was done using the active patches of all training images, and the first $G=10$ principal directions were identified.
        (e) Finally, $1$D PCA expert PDFs were computed similar to the $1$D CNN expert PDFs: for all channels of the last layer of $S_\theta$, for all principal directions, for each training subject. In total, we had $C_L\times G = 160$ PCA expert PDFs for each training subject.
        The hyperparameter, $\lambda$, was empirically set to $0.1$, and $lr_{tta}$ to $0.0001$.
        
    \end{enumerate}
    
    \item \textbf{Analysis Experiments}
    
    \begin{enumerate}[wide, labelwidth=!, labelindent=0pt]
    
        % ================================================================================
        % Approximating Expert Distributions with KDEs rather than as Gaussians
        % ================================================================================
        \item \textbf{Approximating Expert Distributions with KDEs rather than as Gaussians}: In the experiments above, we approximated the individual expert distributions of the FoE model (Eqn.~\ref{eqn_poe_pdf_using_cnn_functions_new_notation}) as Gaussian distributions. As the expert distributions are in $1$D, we also considered non-parametric estimation methods, such as kernel density estimation (KDE)~\cite{scott1977kernel},~\cite{davis2011remarks},~\cite{parzen1962estimation}. This approach is provides an alternative to the \textit{soft-binning}-based non-parametric approximation in \cite{eastwood2022sourcefree}.
        In general, KDEs have the two important downsides. Firstly, the number of data points required to get a reliable density estimate grows exponentially with dimensionality. This is not a concern in low dimensions. Secondly, KDEs require access to the training samples to evaluate the PDF at a given test sample. Again, in low dimensions (e.g 1D), it may be feasible to evaluate and save the KDE over the entire domain of interest when one has access to the training samples. Thus, the training samples are no longer required at test time.
        Accordingly, we compute 
        \begin{equation}~\label{eqn_kde_experts}
          \begin{array}{l}
                p_{cl}(u) = \frac{1}{N_z} \: \sum_{z} \: \frac{1}{N_{xl}*N_{yl}} \sum_{i} \: \frac{1}{\sqrt{2\pi}} \: \exp \: (-\alpha \: ||u - u_i||_2^2))
          \end{array}
        \end{equation}
        Being more expressive than Gaussians, KDEs can potentially capture higher-order moments of the expert distributions - thus leading to more accurate distribution matching and better TTA performance. Implementation-wise, when the $1$D PDFs were estimated as Gaussians, the KL-divergence could be computed in closed form. When KDEs were used, we numerically computed the integral in the KL-divergences using Riemann sums.

        % =====================
        % Effect of the hyper-parameter \lambda
        % =====================
        \item \textbf{Effect of the weighting between the CNN and the PCA experts}: The effect of the weighting parameter, $\lambda$, in Eqn.~\ref{eqn_tta_pca_opt}, was empirically analyzed for the 5 test distributions of the prostate segmentation experiment.

    \end{enumerate}

\end{enumerate}
    
% ================
% KEEP THE RESULTS TABLE HERE FOR BETTER PLACEMENT IN THE PDF
% ================
\begin{table*}[h!]
    \begin{adjustwidth}{-1.5cm}{-1.75cm}
    \centering
    % \footnotesize
    \scriptsize
    % \small
    \begin{tabular}{| c | c|c|c|c|c|c|c|c|c|c|c|c| }
        
        \Xhline{2\arrayrulewidth}
        \rowcolor{titlegray} 
        \backslashbox{Method}{Test}
        % & \scriptsize{BMC}
        % & \scriptsize{USZ}
        % & \scriptsize{UCL}
        % & \scriptsize{HK}
        % & \scriptsize{BIDMC}
        % & \scriptsize{UHE}
        % & \scriptsize{HVHD}
        % & \scriptsize{USZ}
        % & \scriptsize{VU}
        % & \scriptsize{UCL}
        % & \scriptsize{AC}
        % % & \scriptsize{AS}
        % & \scriptsize{NUHS}
        & UCL
        & HK
        & BIDMC
        & BMC
        & USZ
        & UHE
        & HVHD
        & USZ
        & VU
        & UCL
        & AC
        % & AS
        & NUHS
        \\\Xhline{2\arrayrulewidth}
        
        \rowcolor{lightgray}
        & \multicolumn{5}{c|}{{{Prostate}}}
        & \multicolumn{2}{c|}{{{Cardiac}}}
        & \multicolumn{3}{c|}{{{Spine}}}
        & \multicolumn{1}{c|}{{{Brain}}}
        & \multicolumn{1}{c|}{{{WMH}}}
        \\\Xhline{2\arrayrulewidth}
        
        \rowcolor{lightgray} & \multicolumn{12}{c|}{{{Supervised Learning on Training Distribution}}}
        \\\Xhline{2\arrayrulewidth}

        \rowcolor{lightergray} 
        Baseline & % Baseline
        0.50 & % UCL
        0.68 & % HK
        0.29 & % BIDMC
        0.28 & % BMC
        0.67 & % USZ
        0.86 & % UHE
        0.38 & % HVHD
        0.61 & % USZ (site 3)
        0.82 & % VU (site 4)
        0.79 & % UCL (site 1)
        0.69 & % ABIDE-C
        % 0.xxx & % ABIDE-S
        0.00 % NUHS % to be done for different runs
        % \\ \hline
        
        % \rowcolor{lightergray} & \multicolumn{12}{c|}{{{SD Test Set Performance}}} \\ \hline
        
        % \rowcolor{lightergray}
        % & \multicolumn{5}{c|}{{{0.86}}}
        % & \multicolumn{2}{c|}{{{0.82}}}
        % & \multicolumn{3}{c|}{{{0.88}}}
        % & \multicolumn{1}{c|}{{{0.87}}}
        % & \multicolumn{1}{c|}{{{0.71}}}
        \\\Xhline{2\arrayrulewidth}
        
        \rowcolor{lightgray} & \multicolumn{12}{c|}{{{Domain Generalization}}}
        \\\Xhline{2\arrayrulewidth}
        
        \rowcolor{lightergray} 
        Strong baseline \cite{zhang2019generalizing} & % (Strong baseline)
        0.77 & % UCL
        0.82 & % HK
        0.62 & % BIDMC
        0.77 & % BMC
        0.76 & % USZ
        0.85 & % UHE
        0.80 & % HVHD
        0.67 & % USZ (site 3)
        0.84 & % VU (site 4)
        0.88 & % UCL (site 1)
        0.76 & % ABIDE-C
        % 0.xxx & % ABIDE-S
        0.37 % NUHS
        % \\\hline
        
        % \rowcolor{lightergray} & \multicolumn{12}{c|}{{{SD Test Set Performance}}}
        % \\ \hline
        
        % \rowcolor{lightergray}
        % & \multicolumn{5}{c|}{{{0.91}}}
        % & \multicolumn{2}{c|}{{{0.83}}}
        % & \multicolumn{3}{c|}{{{0.89}}}
        % & \multicolumn{1}{c|}{{{0.87}}}
        % & \multicolumn{1}{c|}{{{0.72}}}
        \\\Xhline{2\arrayrulewidth}
        
        \rowcolor{lightgray} & \multicolumn{12}{c|}{{{Test Time Adaptation}}}
        \\\Xhline{2\arrayrulewidth}
        
        \rowcolor{lightergray} 
        Entropy Min. \cite{wang2020tent}
        %, \cite{bateson2020source}\href{https://link.springer.com/content/pdf/10.1007\%2F978-3-030-59710-8_48.pdf}{$^{\uparrow}$}
        & % TTA with entropy minimization
        0.77 & % UCL
        0.81 & % HK
        0.68$^\triangle$ & % BIDMC
        0.77 & % BMC
        0.80$^\blacktriangle$ & % USZ
        0.85 & % UHE
        0.80$^\triangle$ & % HVHD
        0.67 & % USZ (site 3)
        0.84 & % VU (site 4)
        0.88 & % UCL (site 1)
        0.81$^\blacktriangle$ & % ABIDE-C
        % 0.xxx & % ABIDE-S
        0.36$^\blacktriangledown$ % NUHS
        \\\hline
        
        \rowcolor{lightergray} 
        DAE \cite{karani2020test} & % TTA with DAE
        \textbf{0.84}$^\blacktriangle$ & % UCL
        \textbf{0.84}$^\triangle$ & % HK
        \textbf{0.75}$^\blacktriangle$ & % BIDMC
        \textbf{0.81}$^\triangle$ & % BMC
        \textbf{0.82}$^\blacktriangle$ & % USZ
        \textbf{0.87}$^\blacktriangle$ & % UHE
        0.81 & % HVHD
        \textbf{0.69} & % USZ (site 3)
        0.80$^\triangledown$ & % VU (site 4)
        0.80 & % UCL (site 1)
        \textbf{0.82}$^\blacktriangle$ & % ABIDE-C
        % 0.xxx & % ABIDE-S
        0.37 % NUHS
        \\\hline
        
        \rowcolor{lightergray} 
        AE \cite{he2021autoencoder} & % TTA with AE
        0.78 & % UCL
        0.83 & % HK
        0.51$^\triangledown$ & % BIDMC
        \textbf{0.79} & % BMC
        0.79 & % USZ
        0.86$^\blacktriangle$ & % UHE
        0.80 & % HVHD
        \textbf{0.69}$^\triangle$ & % USZ (site 3)
        0.84$^\triangle$ & % VU (site 4)
        0.88$^\triangle$ & % UCL (site 1)
        0.78$^\blacktriangle$ & % ABIDE-C
        % 0.xxx & % ABIDE-S
        0.24$^\blacktriangledown$ % NUHS
        \\\Xhline{2\arrayrulewidth}
        
        \rowcolor{lightergray} 
        FoE-CNN \cite{eastwood2022sourcefree} & % \cite{ishii2021source}\href{https://arxiv.org/pdf/2101.10842.pdf}{$^{\uparrow}$} & % TTA with Gaussian Matching (CNN experts)
        0.78 & % UCL
        0.77$^\triangledown$ & % HK
        0.64 & % BIDMC
        0.76 & % BMC
        0.76 & % USZ
        0.86 & % UHE
        \textbf{0.82}$^\blacktriangle$ & % HVHD
        0.68 & % USZ (site 3)
        \textbf{0.85}$^\triangle$ & % VU (site 4)
        \textbf{0.89}$^\triangle$ & % UCL (site 1)
        0.79$^\blacktriangle$ & % ABIDE-C
        % 0.xxx & % ABIDE-S
        0.24$^\blacktriangledown$ % NUHS
        \\\hline
        
        \rowcolor{lightergray} 
        FoE-CNN-PCA (Ours) & % TTA with Gaussian Matching (CNN and PCA experts)
        0.79 & % UCL
        0.81 & % HK
        0.73$^\triangle$ & % BIDMC
        0.75 & % BMC
        0.78 & % USZ
        0.85 & % UHE
        \textbf{0.82}$^\blacktriangle$ & % HVHD
        0.68 & % USZ (site 3)
        0.83$^\triangledown$ & % VU (site 4)
        0.88 & % UCL (site 1)
        0.79$^\blacktriangle$ & % ABIDE-C
        % 0.xx & % ABIDE-S
        \textbf{0.42}$^\blacktriangle$ % NUHS
        \\\Xhline{2\arrayrulewidth}
        
        \rowcolor{lightgray} & \multicolumn{12}{c|}{{{Transfer Learning}}}
        \\\Xhline{2\arrayrulewidth}
        
        \rowcolor{lightergray} 
        Benchmark & % Benchmark
        0.80 & % UCL
        0.85 & % HK
        0.82 & % BIDMC
        0.83 & % BMC
        0.84 & % USZ
        0.88 & % UHE
        0.83 & % HVHD
        0.78 & % USZ (site 3)
        0.85 & % VU (site 4)
        0.90 & % UCL (site 1)
        0.88 & % ABIDE-C
        % 0.xx & % ABIDE-S
        0.77 % NUHS % to be done for the different runs
        \\\Xhline{2\arrayrulewidth}

    \end{tabular}
    \caption{Dice scores (averaged over all foreground labels and all test subjects) for the segmentation test-distribution datasets. In each column, the highest Dice score among the TTA methods has been highlighted. The Dice scores for test images from the training distribution are: (a) for the baseline: RUNMC 0.86, CSF: 0.82, PM: 0.88, HCP: 0.87, UMC: 0.71, (b) for the strong baseline: RUNMC 0.91, CSF: 0.83, PM: 0.89, HCP: 0.87, UMC: 0.72. Results for the NUHS dataset are mean values over $4$ runs. Paired permutation tests were done to measure the statistical significance of the improvement or degradation caused by each TTA method over the strong baseline. $^\triangle$ ($^\triangledown$) and $^\blacktriangle$ ($^\blacktriangledown$) indicate improvement (degradation) with p-value less than 0.05 and 0.01, respectively. The stricter significance test (p-value 0.01) was done to counter the multiple comparison problem \cite{bland1995multiple}.}~\label{tab:quant_results_segmentation}
    % \vspace{-30pt}
    \end{adjustwidth}
\end{table*}

\subsubsection{Results}
\noindent The following points can be inferred from the quantitative results of our segmentation experiments (Table~\ref{tab:quant_results_segmentation}).
\begin{enumerate}[label=(\roman*), wide, labelwidth=!, labelindent=0pt]

    \item The baseline demonstrates that the DS problem exists for all the 5 anatomies. The difference between the Dice scores on the training and test distributions is sometimes as high as 60 Dice points; a model that provides almost perfect segmentations on test images from the training distribution can potentially provide completely un-usable segmentations on test images from a shifted distribution (e.g. test images from a different hospital).
    
    \item Data augmentation~\cite{zhang2019generalizing} helps vastly. This \textit{strong baseline} is much more robust to DS than the baseline - in some cases, providing a performance jump as high as 50 Dice points. These results corroborate numerous similar findings in the current literature. Given the generality and effectiveness of the approach, we believe it is imperative that works studying DS robustness in CNN-based medical image segmentation should include stacked data augmentation during training.
    
    \item A gap to the benchmark still remains - in most cases, heuristic data augmentation falls short of rivalling the performance of supervised fine-tuning.
    % \item \textbf{Comparison of TTA methods}: For most cases, the performance the TTA methods are similar, with some notable exceptions. Firstly, 
    
    \item Results of the TTA methods are described below. When making statements about statistical significance of results in the text below, we follow a strict threshold based on Benferroni correction to account for the multiple comparison problem~\cite{bland1995multiple}. For each dataset, permutation tests ($n=100000$) were used to compute statistical significance of the performance improvement or degradation provided by each TTA method with respect to the strong baseline. Thus, $5$ comparisons were made for each dataset. So, the p-value threshold was divided by $5$.
    
    \begin{enumerate}[wide, labelwidth=!, labelindent=0pt]
    
        \item Entropy minimization-based TTA~\cite{wang2020tent} does not require construction of additional models to capture training distribution traits; yet, it provides performance improvement in some cases. Also, unlike other works~\cite{bateson2020source}, we largely do not observe the problem that the entropy minimization leads to all pixels being predicted as the same class. This might have been due to the limited adaptation ability provided by $N_\phi$.
        
        \vspace{2pt} \noindent However, the performance gains are statistically significant for only 2 out of the 12 test datasets. As well, the performance degrades the strong baseline significantly (although marginally) for the lesion test dataset. Another downside of this method is that it can only be applied for tasks with categorical outputs.
        
        \item TTA-DAE~\cite{karani2020test} provides the best performance for the most number of test datasets for healthy tissue segmentations. For 5 out of 11 healthy test datasets, the improvements provided by this method over the strong baseline are statistically significant.
        It also leads to a drop of $5$ and $8$ Dice points in the mean results for the two spine datasets; however, permutation tests show that the drops may be due to large degradation for a small number of test subjects within those datasets. Even so, the large drops in performance for particular subjects may be indicative of the DAE's DS problem - that is, the DAE's outputs may be unreliable when it is fed with segmentations that do not match the heuristically designed noise distribution used for its training.
        
        \vspace{2pt} \noindent Furthermore, the DAE fails to improve performance for the lesion dataset. We believe that this reflects its inapplicability to tackle the DS problem in anatomies where reliable shape priors cannot be learned.
        
        \vspace{2pt} \noindent In terms of applicability, the DAE-based TTA is also restricted in terms of the tasks that it can applied to. For segmentation, the DAE could be trained by heuristically designing a suitable corruption distribution. It is unclear how to achieve this for other tasks. 
        
        \item Autoencoder-based TTA~\cite{he2021autoencoder} provides performance improvement in several cases. However, statistically significant improvements could only be obtained for $2$ test datasets; even in these cases, the improvements were marginal ($1$ and $2$ dice points). This method also also lead to a drop of $12$ and $13$ Dice points for the prostate BIDMC and the WMH dataset, respectively; the latter was statistically significant.
    
        \item The FoE-CNN that in principle resembles the concurrent approach of \cite{eastwood2022sourcefree} is overall, less performant than the PCA-based extended FoE proposed in the current work.
    
        \item As compared to the strong baseline, the proposed FoE-CNN-PCA based TTA improves performance for $7$ and retains performance for $2$ out of the $12$ test distributions. In particular, the proposed method shows promising performance gains in cases where the other task-agnostic methods falter substantially (e.g. prostate BIDMC and WMH). Out of these, the improvements are statistically significant for 3 test datasets, including the lesion dataset.
        
        \vspace{2pt} \noindent The $3$ test distributions where the method leads to a performance drop, the drop is relatively small: $3$, $1$ and $1$ Dice points. We claim that this illustrates the stability of the proposed TTA method and validates our initial hypothesis - FoE-based TTA improves performance in the face of acquisition-related DS in medical imaging, while itself being substantially more robust to the DS shift problem that other priors such as the DAE~\cite{karani2020test} or the AE~\cite{he2021autoencoder} may be vulnerable to.
        
        \vspace{2pt} \noindent Importantly, the proposed method provides the best performance for the task of WMH segmentation - indicating its superiority in cases where CNN-based helper modules such as DAEs~\cite{karani2020test} may be unable to learn appropriate shape priors. Notably, all competing methods from the literature fail to improve DS robustness for this lesion segmentation experiment; the proposed method is the only approach that shows promising results in this challenging scenario.
    
    \end{enumerate}
    
    \item Analysis Experiments
    \begin{enumerate}[wide, labelwidth=!, labelindent=0pt]
    \item Approximating Expert Distributions with KDEs rather than as Gaussians: Comparing the KDEs v/s Gaussian approximations (Fig.~\ref{fig_gaussian_vs_kde}), we observed that the actual distributions do not differ substantially from their Gaussian approximations. This is also reflected in the TTA results in Table~\ref{table_tta_gaussian_vs_kde} - performance of the proposed method is very similar for both estimates of expert distributions.
    \item Effect of the weighting between the CNN and the PCA experts: Results of this hyper-parameter tuning are shown in Table~\ref{table_tta_pca_lambda}. The introduction of PCA experts with $\lambda = 0.1$ improves TTA performance for 4 of the 5 prostate datasets. However, increasing $\lambda$ to $1.0$ leads to performance decrease in 3 of the 5 datasets. Based on these results, we choose $\lambda=0.1$ for all datasets of all anatomies.
    \end{enumerate}

\end{enumerate}

% ================
% Comparing Gaussing vs KDEs
% ================
\begin{figure}[h!]
\centering
    \includegraphics[trim = 0mm 230mm 0mm 0mm, angle=0, clip, width=0.7\textwidth]{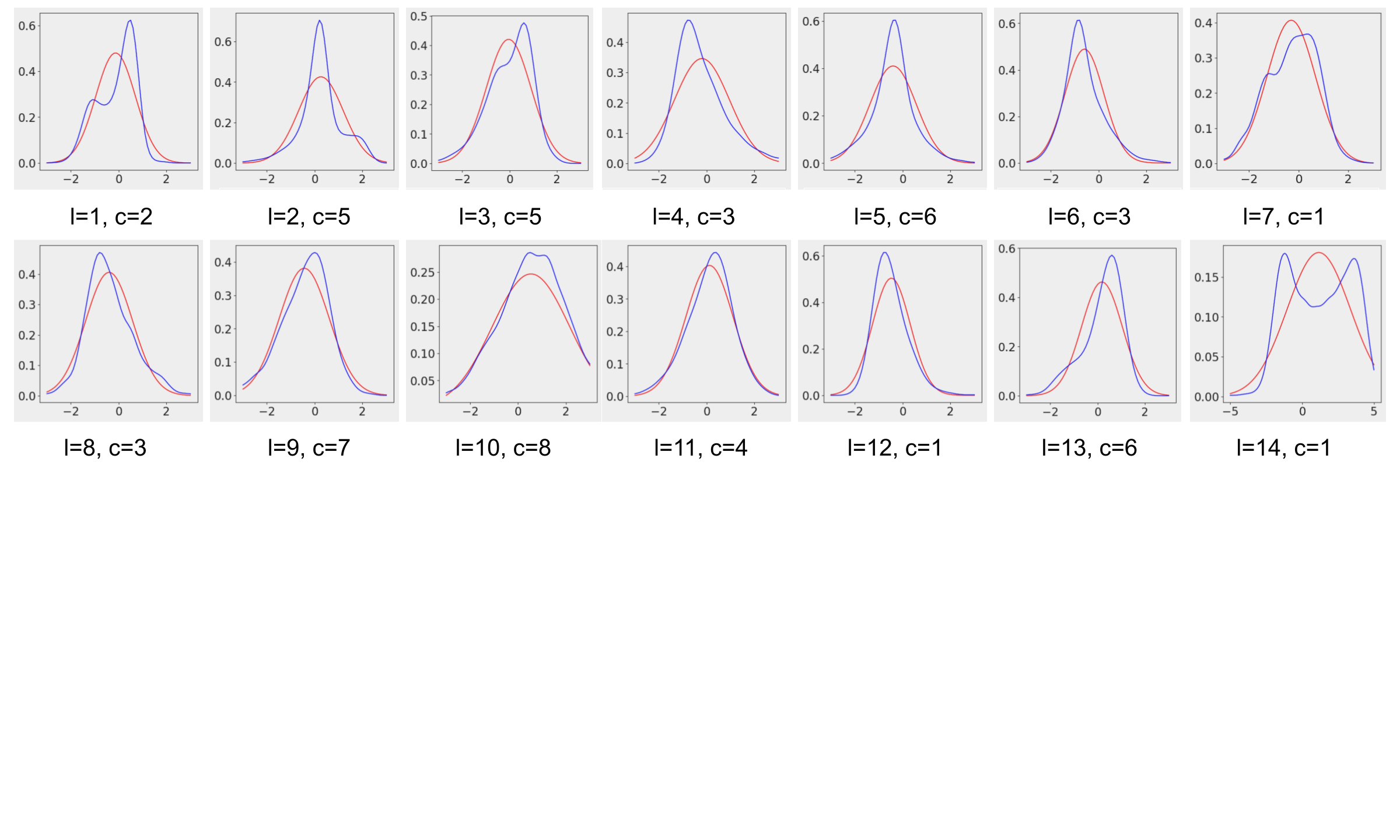}
\caption[TTA-FoE - KDE v/s Gaussian approximation of task-specific expert distributions]{Comparison of KDEs v/s Gaussian approximations (corresponding to a single prostate RUNMC training subject) for modeling the channel PDFs of different layers of the trained segmentation network. $l=14$ is the last-but-one layer of the network. From here, a $1$x$1$ convolution gives the segmentation logits. In each layer ($l$), the channel ($c$) with the visually most-non-gaussian KDE is chosen for visualization. With this choice, some non-Gaussianity is observed in the initial and final layers, while the layers in the middle of the segmentation CNN has highly Gaussian marginal distributions.}~\label{fig_gaussian_vs_kde}
\vspace{-10pt}
\end{figure}

\begin{table}[h!]
    \centering
    \small
    \begin{tabular}{ c c c c c c }
        
        \hline
        \rowcolor{titlegray} 
        \backslashbox{Method}{Test}
        & UCL & HK & BIDMC & BMC & USZ \\ \hline

        \rowcolor{lightgray} & \multicolumn{5}{c}{{{TTA-FoE-CNN-PCA}}}
        \\\hline
        \rowcolor{lightergray} Gaussian
        & 0.79 % UCL 
        & 0.81 % HK
        & 0.73 % BIDMC 
        & 0.75 % BMC 
        & 0.78 % USZ 
        \\ % \hline
        
        \rowcolor{lightergray} KDE
        & 0.79 % UCL 
        & 0.81 % HK
        & 0.74 % BIDMC 
        & 0.76 % BMC 
        & 0.78 % USZ 
        \\ \hline
        
    \end{tabular}
    \caption[TTA-FoE - KDE vs Gaussian approximation of FoE experts]{Effect of approximating $1$D distributions of the FoE model with Gaussians v/s kernel density estimation (KDE). Both approximations lead to very similar TTA performance. Fig.~\ref{fig_gaussian_vs_kde} provides visual justification of this observation - the 1D distributions of CNN as well as the PCA experts are sufficiently well approximated with Gaussians.}~\label{table_tta_gaussian_vs_kde}
    \vspace{-20pt}
\end{table}

% =====================
% Effect of the hyper-parameter \lambda
% =====================
\begin{table}[h!]
    \centering
    \small
    \begin{tabular}{ c c c c c c }
        
        \hline
        \rowcolor{titlegray} 
        \backslashbox{Method}{Test}
        & UCL & HK & BIDMC & BMC & USZ \\\hline

        \rowcolor{lightgray} & \multicolumn{5}{c}{{{TTA-FoE-CNN}}}
        \\\hline
        \rowcolor{lightergray} $\lambda = 0.0$
        & 0.78
        & 0.77
        & 0.64
        & 0.76
        & 0.76
        \\\hline
        
        \rowcolor{lightgray} & \multicolumn{5}{c}{{{TTA-FoE-CNN-PCA}}}
        \\\hline
        \rowcolor{lightergray} $\lambda = 0.1$
        & \textbf{0.79}
        & 0.81
        & 0.73
        & 0.75
        & \textbf{0.78}
        \\ % \hline
        
        \rowcolor{lightergray} $\lambda = 1.0$
        & 0.77
        & \textbf{0.82}
        & \textbf{0.74}
        & 0.74
        & 0.77
        \\ \hline
        
    \end{tabular}
    \caption[TTA-FoE - Effect of weighting parameter for the CNN and PCA experts]{Effect of the weighting parameter between the CNN and PCA experts in TTA-FoE-CNN-PCA. Based on these results, we choose $\lambda=0.1$ for all datasets of all anatomies.}~\label{table_tta_pca_lambda}
    % \vspace{-25pt}
\end{table}

% =========================================
% REGISTRATION
% =========================================
\newpage
\subsection{Registration}~\label{sec_registration}
\noindent Next, we checked if the proposed method can tackle acquisition-related DS in another task of high practical importance - registration of brain scans with an atlas.
The registration CNN is set up as follows. \footnote{Ideally, such registration would be done in $3$D. However, to avoid memory issues in $3$D CNNs, we conduct experiments in a $2$D setup. We believe that this still serves as credible evidence of the method's applicability in this task.} Let $a$ be an atlas and $x$ be the image. Let $a_s$ and $x_s$ be the corresponding segmentation labels. We treat $a$ as the moving image and register it to $x$, the fixed image. $x$ is first passed through the normalization module, $N_\phi$, to obtain $z$. $z$ and $a$ are concatenated and passed through a deep CNN, $S_\theta$, which outputs a velocity field $v_0$. $v_0$ is exponentiated via a \textit{squaring-and-scaling} layer~\cite{dalca2018unsupervised} to obtain a diffeomorphic deformation field, $\Phi$. The Dice loss between the warped moving segmentation, $a_s \odot \Phi$, and $x_s$ is used for training $N_\phi$ and $S_\theta$. For each test image, $N_\phi$ is adapted with the proposed TTA method.

% ================
% DATASETS
% ================
\vspace{5pt} \subsubsection{Datasets}
\begin{figure}[b!]
\centering
    \includegraphics[trim = 5mm 5mm 5mm 5mm, angle=0, clip, width=0.117\textwidth]{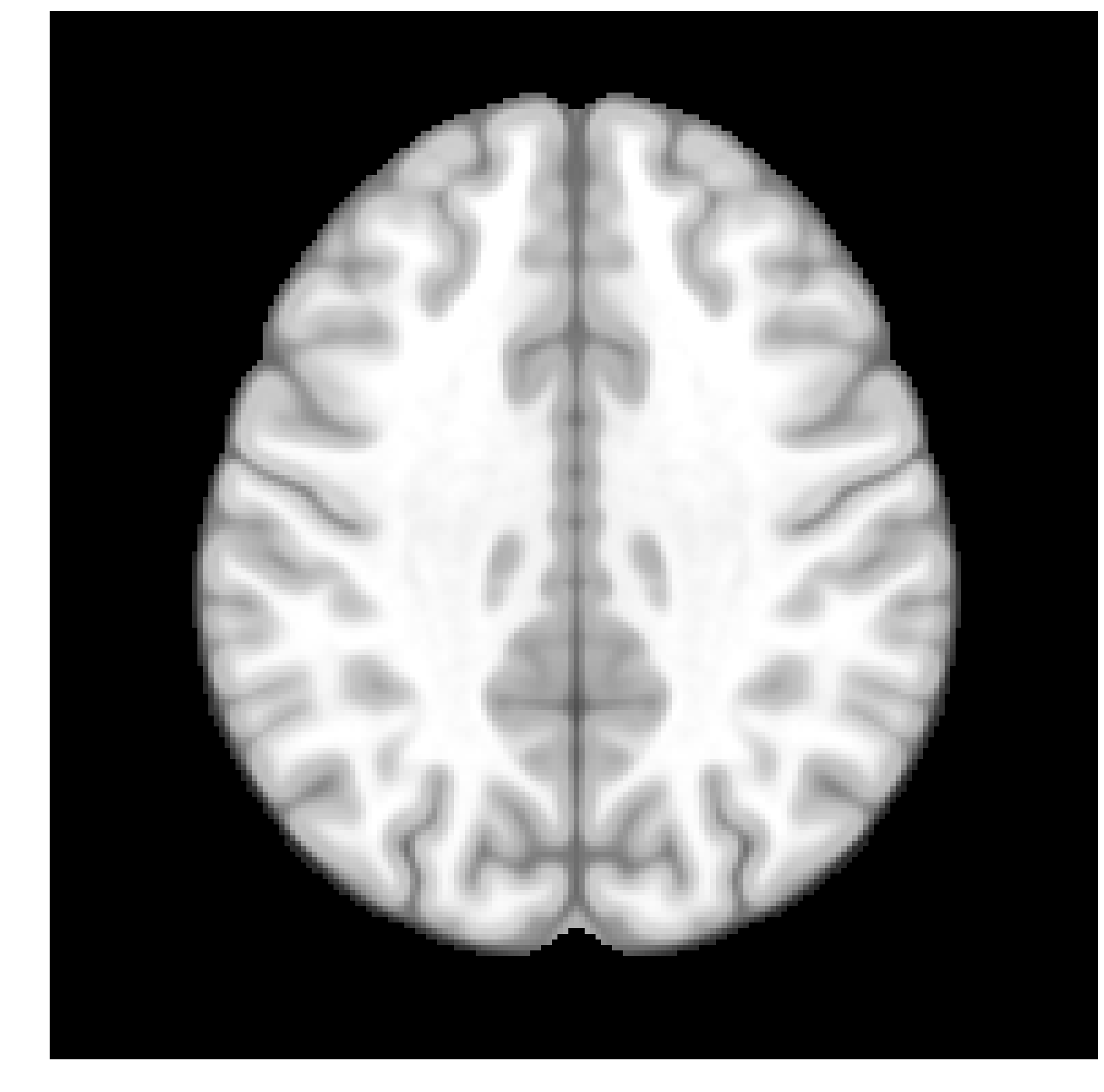}
    \includegraphics[trim = 5mm 5mm 5mm 5mm, angle=0, clip, width=0.117\textwidth]{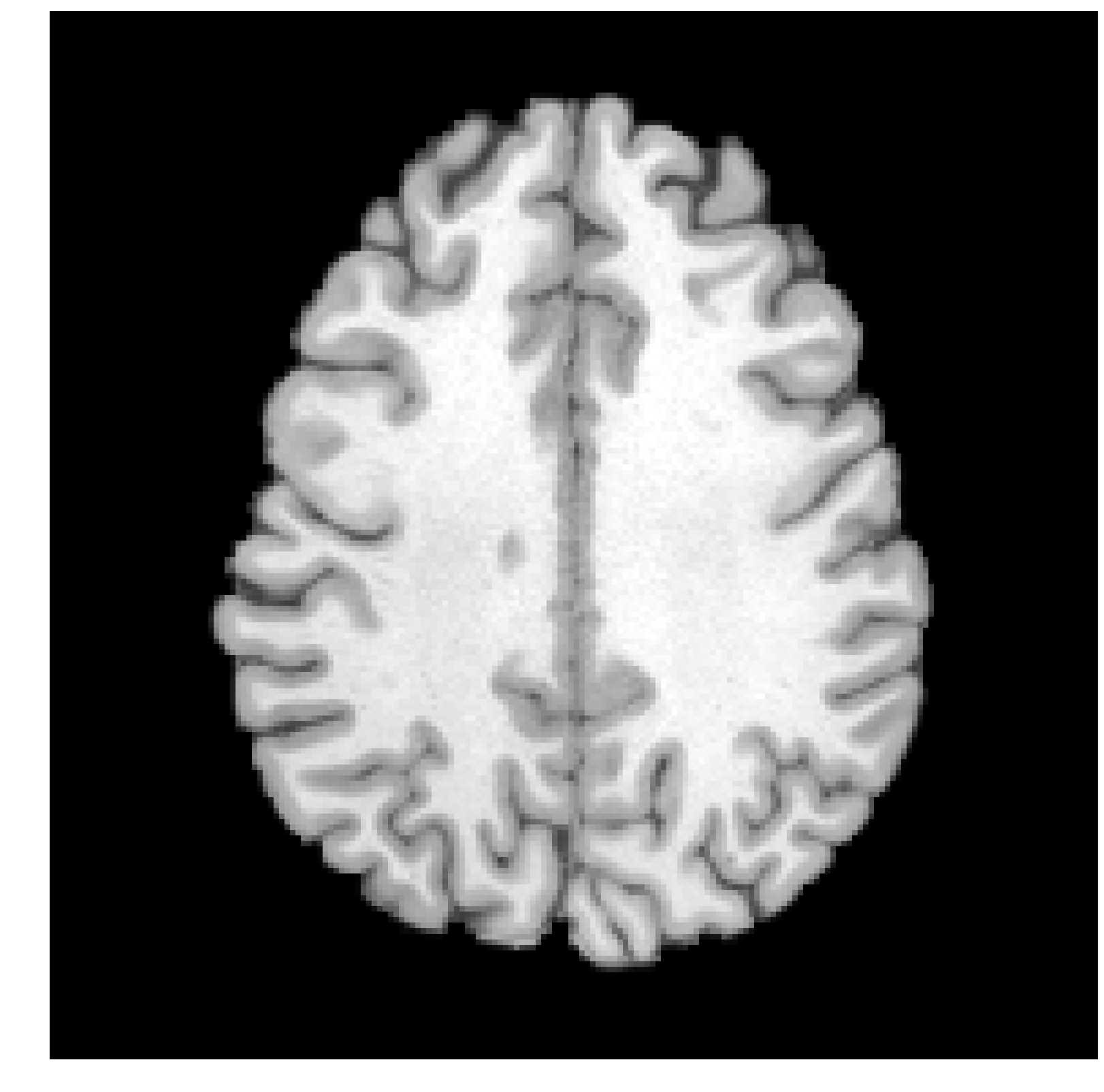}
    \includegraphics[trim = 5mm 5mm 5mm 5mm, angle=0, clip, width=0.117\textwidth]{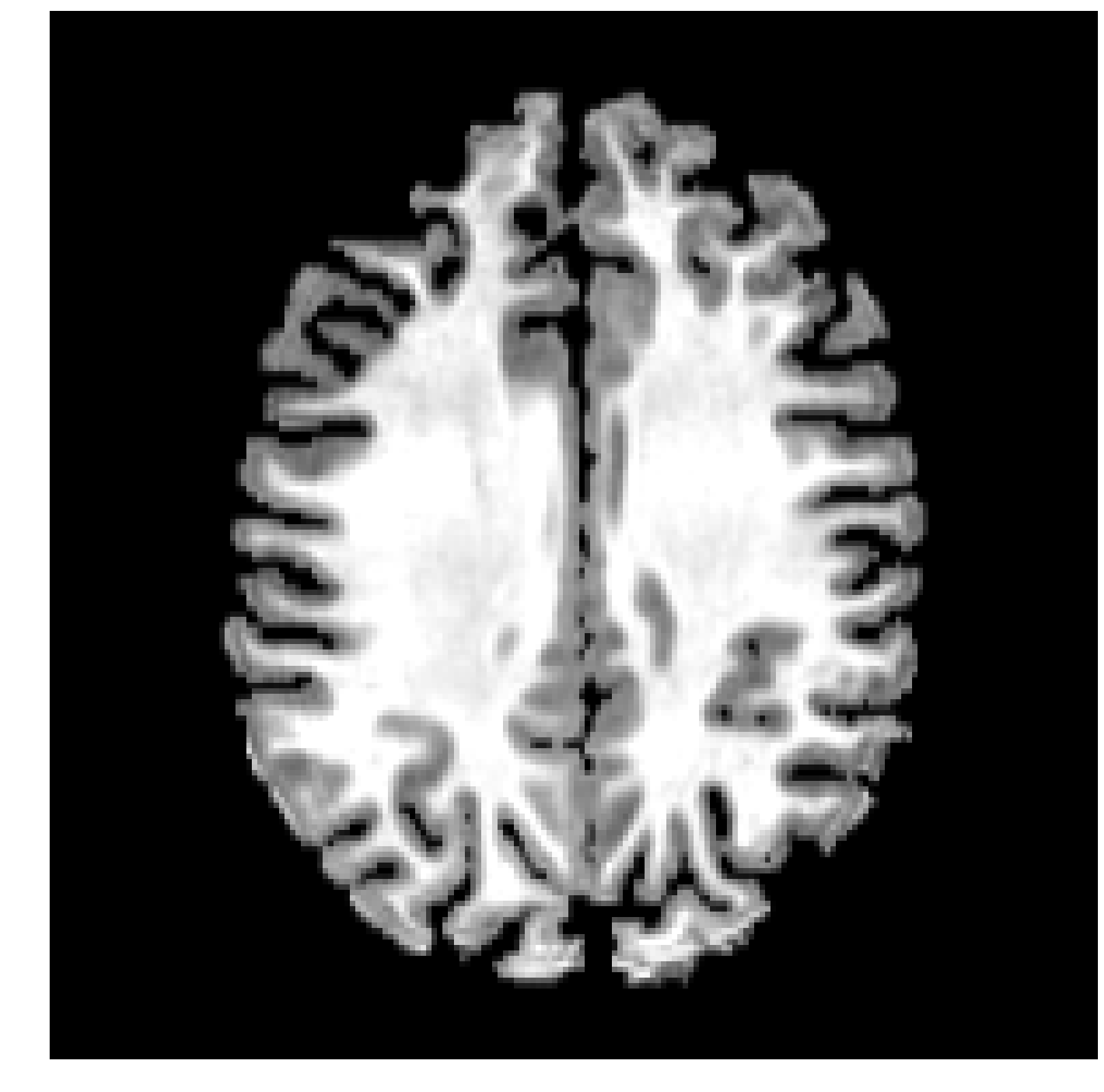}
    \includegraphics[trim = 5mm 37mm 5mm 37mm, angle=0, clip, width=0.117\textwidth]{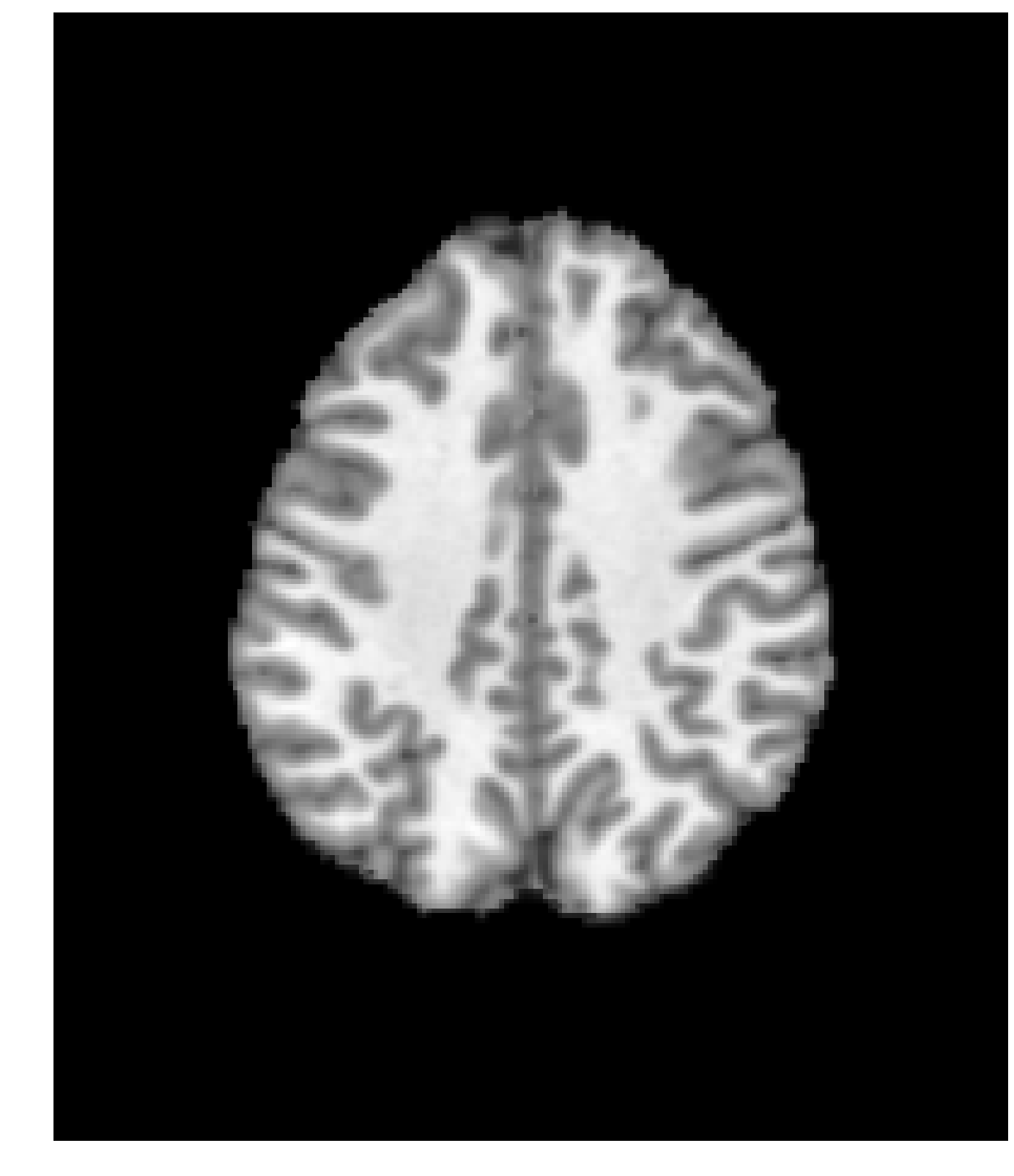}
    
\caption{From left to right: a $2$D slice from the atlas and example slices from three datasets: HCP, ABIDE-STANFORD (AS) and OASIS.}~\label{fig_dataset_details_reg}
\end{figure}

\noindent We used HCP \cite{van2013wu} T1w images as those from the training distribution and ABIDE-STANFORD (AS) \cite{di2014autism} and OASIS \cite{daniel2007oasis} as two test distributions.
We used the atlas provided by \cite{fonov2011unbiased}.
Example images are shown in Fig.~\ref{fig_dataset_details_reg}.
% ================
% Example images
% ================

% ================
% IMPLEMENTATION DETAILS
% ================
\vspace{5pt} \subsubsection{Implementation details}
\noindent All images were re-sampled to an isotropic $1 \: mm^3$ resolution.
Upon visual inspection, the axial slices of the atlas, the HCP and OASIS datasets were roughly aligned in the through plane direction, while the AS volumes were shifted by $10$ slices.
After accounting for this, we extracted the central $40$ axial slices from all volumes. We used 3-label (background, white matter, grey matter) Freesurfer~\cite{fischl2012freesurfer} segmentations for HCP, AS and expert segmentations for the atlas and OASIS.

\vspace{2pt} \noindent Among the TTA methods, we note that TTA-Entropy-Min. \cite{wang2020tent} can only be applied in cases where $S_\theta$ outputs a probability distribution over a fixed number of classes; it is unclear how to extend this for regression. Also, TTA-DAE \cite{karani2020test} requires a denoising autoencoder to be trained with corruption patterns that are expected at test time. Designing such corruptions for the registration task is non-trivial. Thus, we compare the proposed method with TTA-AE~\cite{he2021autoencoder} only.

\begin{table}[h!]
    \centering
    % \footnotesize
    % \scriptsize
    \small
    \begin{tabular}{| c | c|c|c| }
        
        \hline
        \rowcolor{titlegray} 
        \backslashbox{Method}{Test} & HCP & AS & OASIS \\ \hline
        
        \rowcolor{lightergray} Baseline & 0.847 & 0.751 & 0.864 \\ \hline
        
        \rowcolor{lightergray} Strong baseline \cite{zhang2019generalizing} & 0.843 & 0.786 & 0.873 \\ \hline

        \rowcolor{lightergray} TTA-AE~\cite{he2021autoencoder} & - & 0.795 & 0.868 \\ \hline
        
        \rowcolor{lightergray} TTA-FoE-CNN-PCA & - & 0.795$^\triangle$ & 0.870$^\triangledown$ \\ \hline
        
        \rowcolor{lightergray} Benchmark & - & 0.821 & 0.883 \\ \hline

    \end{tabular}
    \caption{Dice scores (averaged over all foreground labels and all test subjects) for the registration experiments. We measured the statistical significance of the improvement or degradation caused by each TTA method over the strong baseline, using paired permutation tests. $^\triangle$ ($^\triangledown$) and $^\blacktriangle$ ($^\blacktriangledown$) indicate improvement (degradation) with p-value less than 0.05 and 0.025, respectively. The stricter significance test (p-value 0.025) was done to counter the multiple comparison problem \cite{bland1995multiple}.}~\label{tab:quant_results_registration}
    \vspace{-20pt}
\end{table}

% ================
% RESULTS
% ================
\vspace{5pt} \subsubsection{Results}
\begin{figure}[t!]
\centering
    \includegraphics[trim = 0mm 270mm 230mm 0mm, angle=0, clip, width=0.7\textwidth]{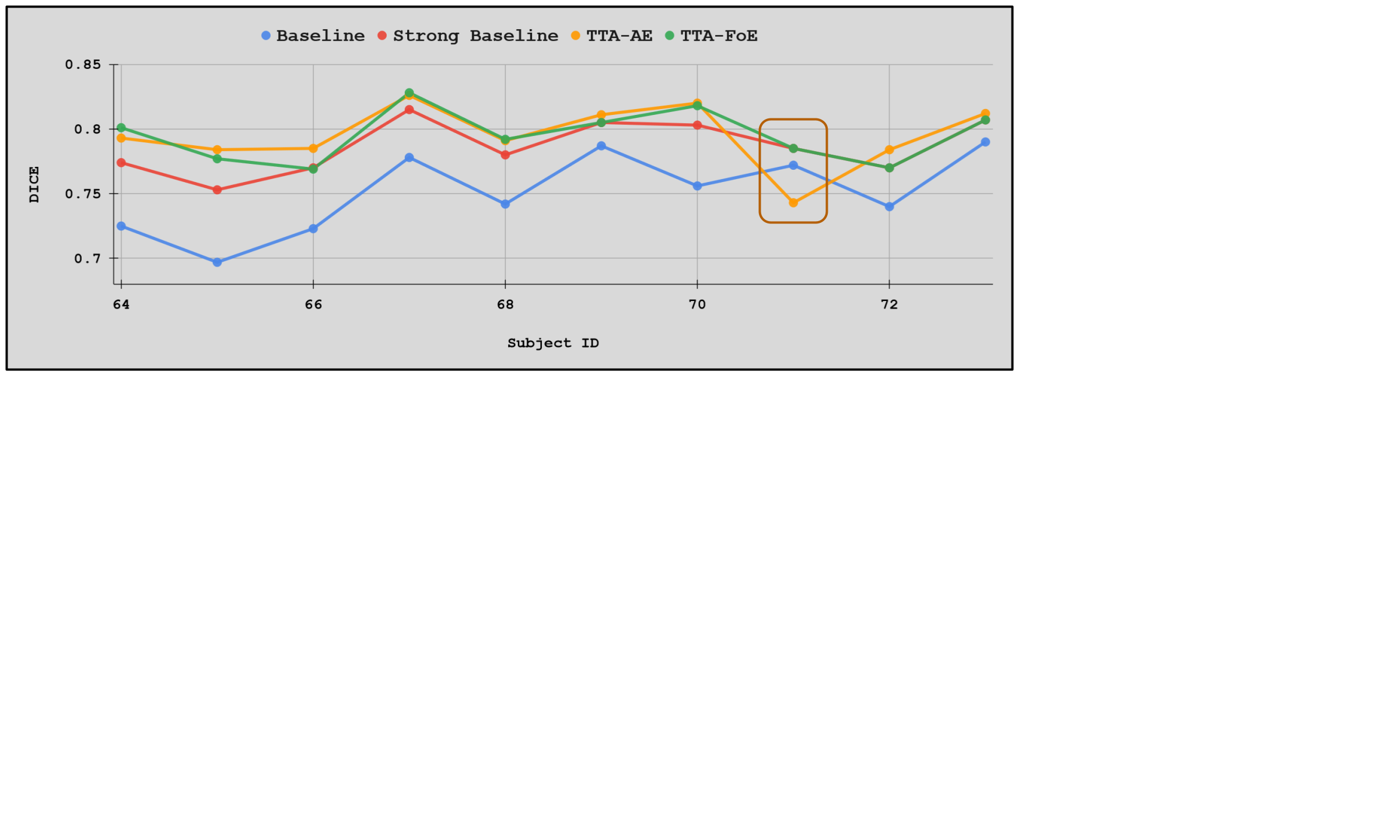}
\caption[TTA-FoE - Subject-wise registration results]{Dice scores for individual subjects of the AS dataset. Overall, both TTA methods perform similarly well for most subjects. However, the brown box highlights one subject, where TTA-AE leads to performance degradation as compared to both the baseline as well as the strong baseline. Such degradation does not occur with the proposed method.}~\label{fig_reg_subjectwise}
\end{figure}

\noindent The following points can be inferred from the quantitative results of our registration experiments (Table~\ref{tab:quant_results_registration}).
\begin{enumerate}[label=(\roman*), wide, labelwidth=!, labelindent=0pt]
    \item In the baseline, the DS problem is quite stark for the AS dataset, but relatively mild for the OASIS dataset.
    \item Stacked data augmentation~\cite{zhang2019generalizing} (with the same hyperparameters as in the segmentation experiments) provides substantial gains for registration as well. In this \textit{strong baseline}, the performance of OASIS is already almost as good as in the benchmark. However, a gap between the \textit{strong baseline} and the benchmark exists for the AS dataset. In a practical setting, TTA methods would typically be unaware of the extent to which the DS problem exists before TTA. In this scenario, TTA methods should ideally improve performance for datasets which suffer from the DS problem, and retain performance for other datasets.
    \item On average, both the proposed method and TTA-AE~\cite{he2021autoencoder} improve the performance of the strong baseline for the AS dataset and retain it for the OASIS dataset. However, the changes in performance brought about by TTA are not statistically significant for any dataset. Subject-wise results are shown in Fig.~\ref{fig_reg_subjectwise}.
    
    \vspace{2pt} \noindent We believe that this set of experiments demonstrates that, in principle, the proposed method can be applied to the image registration task. Further, it achieves comparable results in this task to a previously existing task-agnostic TTA method~\cite{he2021autoencoder}. However, the registration task seems to be particularly challenging for both methods.
\end{enumerate}

% ================
% Comparing Gaussing vs KDEs
% ================
\section{Discussion}~\label{sec_discuss}
In this section, we discuss the strengths and limitations of the proposed methods, and outline avenues for further research.

\subsection{Strengths of the proposed method}
\begin{enumerate}[wide, labelwidth=!, labelindent=0pt]

    % Only method to provide improvement for lesion datasets. indicates that the proposed method is more robust that other TTA methods for this scenario.
    \item \textbf{Lesion segmentation performance}: All existing TTA methods failed to tackle the DS robustness problem for lesion datasets. Furthermore, $3$ out of the $4$ existing methods lead to statistically significant performance degradation over the strong baseline. In particular, TTA-DAE, which shows strong performance for healthy tissue segmentation, fails to improve performance for lesions due to the difficulty in learning appropriate shape priors. The proposed method provided substantial as well as statistical significant performance improvement in this challenging scenario.
    
    % Task-agnostic. in principle, can be applied at least!
    \item \textbf{Applicability to multiple tasks}: Our experiments indicate that the proposed method can, in principle, be applied to multiple tasks. Such generality is an important asset; the DS problem is likely to occur in all medical image analysis tasks.
    
    % General formulation
    \item \textbf{Generalization of previous works}: This work makes the novel contribution of casting the marginal distribution matching idea in a Field-of-Experts formulation. This observation allows us view several recent works \cite{li2018adaptive},~\cite{schneider2020improving},~\cite{ishii2021source},~\cite{eastwood2022sourcefree} as instances of our general framework, and enables us to build on these works by introducing additional expert functions in the form of principle loadings of feature patches.

\end{enumerate}

\subsection{Limitations of the proposed method}

\begin{enumerate}[wide, labelwidth=!, labelindent=0pt]

    \item \textbf{Performance on healthy tissue segmentation is not as good as TTA-DAE}: Although the proposed method improves performance of the strong baseline in a large number of the test datasets, methods specifically designed for image segmentation often outperform the more general method developed in this work.
    
    \item \textbf{Matching the distribution of individual experts rather than the full FoE distribution}:
    An important relaxation in TTA-FoE is between Eqn.~\ref{eqn_tta_poe_full} and Eqn.~\ref{eqn_tta_opt}. Eqn.~\ref{eqn_tta_poe_full} seeks to match the full FoE distribution between test and training images. However, this is not possible as the computation of the normalization constant $\mathcal{C}$ is intractable. Thus, we instead carry out the relaxed optimization, as shown in Eqn.~\ref{eqn_tta_opt} - minimizing divergence between the distributions of individual experts. It is unclear if the relaxed optimization is theoretically guaranteed to converge, or if the alignment of individual experts may compete with one another. In practice, we observe the optimization to converge for all the test images, across all test distributions and anatomical regions. We believe that this behaviour could have been aided by the initial closeness of the individual expert distributions. Thus, the proposed TTA method works well for \textit{small DS} (due to changing scanners or acquisition protocol parameters within the same imaging modality), but may not be suitable for \textit{large DS} (for instance, across imaging modalities).
    
\end{enumerate}
    
\subsection{Avenues for further exploration}

\begin{enumerate}[wide, labelwidth=!, labelindent=0pt]

    \item \textbf{Choice of expert functions of the FoE Model}:
    % In the product-of-experts \cite{hinton2002training}, \cite{welling2002learning} or field-of-experts \cite{roth2005fields} formulation, the probability density function (PDF) of a high dimensional random variable is expressed as a normalized product of PDFs of multiple lower-dimensional projections of the random variable. The projection functions are called the experts or the expert functions.
    In initial product-of-experts \cite{hinton2002training}, \cite{welling2002learning} and field-of-experts \cite{roth2005fields} works, the experts are parameterized and learned from data, such that the probability model assigns high likelihood values to the true data - for example, using algorithms such as contrastive divergence. Further, parameters of the expert PDFs are also learned from data.
    In contrast, in this work, we used two types of experts - (1) the task-specific convolutional filters learned in the segmentation or registration CNN and (2) projections onto principal components of patches in the last layer of the segmentation or registration CNN. Thus, we used task-specific experts, and only learned the parameters of the expert PDFs from data.
    In other words, we aligned the test and training normalized images, in terms of their projections that are the most relevant for the task CNN to perform the task at hand. Such a task-specific probability model could be augmented with learned experts, as proposed in earlier works \cite{hinton2002training}, \cite{welling2002learning}, \cite{roth2005fields}. The extended model would potentially capture further projections of the normalized images, apart from the task-specific projections considered in this work. It is unclear if alignment along such directions between test and training images would further improve TTA performance; we defer this analysis to future work.
    
    \item \textbf{Choice of the divergence measure to be minimized for TTA}: We minimize the KL-divergence between expert distributions. It may be interesting to investigate if aligning distributions by minimizing other divergences may lead to improved TTA performance. For instance, in concurrent work, \cite{eastwood2022sourcefree} minimize a symmetric version of the KL divergence. Leveraging the low-dimensionality of the expert outputs, even divergence measures that cannot be computed in closed form, may be easy to compute numerically.

\end{enumerate}

\section*{Acknowledgments}
This work was supported by the following grants:
(a) Swiss Platform for Advanced Scientific Computing (PASC),
(b) Swiss National Science Foundation Grant $205320$-$200877$,
(c) Clinical Research Priority Program (CRPP) Grant on Artificial Intelligence in Oncological Imaging Network, University of Zurich.

\bibliographystyle{IEEEbib}
\bibliography{ref}

\newpage
% ===============================================
% ===============================================
\newpage
\section{Appendix}

% =====================
% Approximating KL-divergence minimization of PoEs / FoEs with KL-divergence minimization of individual experts
% =====================
\subsection[Matching Full FoE Distribution v/s Matching Individual Expert Distributions]{Approximating KL-divergence minimization of the full FoE model with KL-divergence minimization of individual expert distributions}~\label{appendix_approximate_with_individual_KLs}
We show this analysis for Product of Experts (PoEs).
It also holds for FoEs, which are a specific instance of the PoEs formulation.
Consider PoE models for the source and target domain normalized images.
\begin{equation*}
    p^s(z) = \frac{\hat{p}^s(z)}{\mathcal{C}_s}, 
    \: \: \: \mathcal{C}_s = \int_z \hat{p}^s(z) dz,
    \: \: \: \hat{p}^s(z) = \prod_{j=1}^{J} p_j^s(u_j),
    \: \: \: u_j = f_j(z)
\end{equation*}
\begin{equation*}
    p^t(z) = \frac{\hat{p}^t(z)}{\mathcal{C}_t}, 
    \: \: \: \mathcal{C}_t = \int_z \hat{p}^t(z) dz,
    \: \: \: \hat{p}^t(z) = \prod_{j=1}^{J} p_j^t(u_j),
    \: \: \: u_j = f_j(z)
\end{equation*}

\noindent Here, we explicitly show the subscript $j$ in variables $u$ to indicate that different experts have different $1$D co-domains. Now, consider KL-divergence minimization between these distributions:
\begin{equation*}
    min_\phi \: D_{KL} (p^s(z), p^t(z)) \rightarrow min_\phi \: \int_z \: p^s(z) \: \log \: \frac{p^s(z)}{p^t(z)} \: dz
\end{equation*}
\begin{equation*}
    \rightarrow min_\phi \: \int_z \: \frac{\hat{p}^s(z)}{\mathcal{C}_s} \: \log \: \frac{\mathcal{C}_t}{\mathcal{C}_s} \: \frac{\hat{p}^s(z)}{\hat{p}^t(z)} \: dz
\end{equation*}
\begin{equation*}
    \rightarrow min_\phi \: \int_z \: \frac{\hat{p}^s(z)}{\mathcal{C}_s} \: \log \: \frac{\mathcal{C}_t}{\mathcal{C}_s} \: dz + \: \int_z \: \frac{\hat{p}^s(z)}{\mathcal{C}_s} \: \frac{\hat{p}^s(z)}{\hat{p}^t(z)} \: dz
\end{equation*}
% As $\mathcal{C}_s$ and $\mathcal{C}_t$ do not depend on $z$ (it is integrated out), we have:
% \begin{equation*}
%     = min_\phi \: \frac{1}{\mathcal{C}_s} \: \log \: \frac{\mathcal{C}_t}{\mathcal{C}_s} \int_z \: \hat{p}^s(z) \: dz + \: \int_z \: \frac{\hat{p}^s(z)}{\mathcal{C}_s} \: \frac{\hat{p}^s(z)}{\hat{p}^t(z)} \: dz
% \end{equation*}
\begin{equation*}
    \rightarrow min_\phi \: \log \: \frac{\mathcal{C}_t}{\mathcal{C}_s} + \: \int_z \: \frac{\hat{p}^s(z)}{\mathcal{C}_s} \: \frac{\hat{p}^s(z)}{\hat{p}^t(z)} \: dz
\end{equation*}

\noindent Note that during TTA, $\phi$ is fixed for computing the source-domain distribution, while is variable for computing the target-domain distribution.
Thus, ignoring the 'source-domain-only' terms, the minimization can be stated as follows:
\begin{equation*}
    \rightarrow min_\phi \: \log \: \mathcal{C}_t + \int_z \: \hat{p}^s(z) \: \log \: \frac{\hat{p}^s(z)}{\hat{p}^t(z)} \: dz
\end{equation*}

\begin{equation*}
    \begin{array}{l}
        \rightarrow min_\phi \: \log \: \mathcal{C}_t + \\
        \int_{u_1,u_2,..u_J} \: \prod_{j=1}^{J} p_j^s(u_j) \: \log \: \frac{\prod_{j=1}^{J} p_j^s(u_j)}{\prod_{j=1}^{J} p_j^t(u_j)} \: du_1du_2...du_J
    \end{array}
\end{equation*}

\begin{equation*}
    \begin{array}{l}
    \rightarrow min_\phi \: \log \: \mathcal{C}_t + \\
    \sum_{j=1}^{J} \: \int_{u_1,u_2,..u_J} \: \prod_{j=1}^{J} p_j^s(u_j) \: \log \: \frac{p_j^s(u_j)}{p_j^t(u_j)} \: du_1du_2...du_J
    \end{array}
\end{equation*}

\begin{equation*}
    \rightarrow min_\phi \: \log \: \mathcal{C}_t + \sum_{j=1}^{J} \: \int_{u_j} \: p_j^s(u_j) \: \log \: \frac{p_j^s(u_j)}{p_j^t(u_j)} \: du_j
\end{equation*}

\noindent As the normalization constant $\mathcal{C}_t$ is intractable, we ignore it in our optimization:
\begin{equation*}
    \approx min_\phi \: \sum_{j=1}^{J} \: \int_{u_j} \: p_j^s(u_j) \: \log \: \frac{p_j^s(u_j)}{p_j^t(u_j)} \: du_j
\end{equation*}
\begin{equation*}
    \rightarrow min_\phi \: \sum_{j=1}^{J} \: D_{KL}(p_j^s(u_j), p_j^t(u_j))
\end{equation*}

% =====================
% How to incorporate information from multiple training subjects?
% =====================
\subsection[Incorporating Information from Different Training Images]{How to incorporate information from multiple training subjects?}~\label{appendix_multiple_training_subjects}
Consider the KL-divergence between the expected distribution over all training subjects and the distribution of the test subject. For simplicity of notation, let us consider only one $1$D expert's distribution.
\begin{equation*}
    D_{KL}\big{(}E_{p(s)}[p^s(u)], p^t(u)\big{)}
\end{equation*}
\begin{equation*}
    = \int_u \Big{(} \int_s p(s) p^s(u) ds \Big{)} \: log \: \frac{\int_s p(s) p^s(u) ds}{p^t(u)} \: du
\end{equation*}
\begin{equation*}
    = \int_s p(s) \Big{(} \int_u p^s(u) \: log \: \frac{\int_s p(s) p^s(u) ds}{p^t(u)} \: du \Big{)} ds
\end{equation*}
\begin{equation*}
    = \int_s p(s) \Big{(} \int_u p^s(u) \: log \: \frac{\int_s p(s) p^s(u) ds}{p^t(u)} \: \frac{p^s(u)}{p^s(u)} \: du \Big{)} ds
\end{equation*}
\begin{equation*}
    = \int_s p(s) \Big{(} \int_u p^s(u) \: log \: \frac{\int_s p(s) p^s(u) ds}{p^s(u)} \: du + \: \int_u p^s(u) \: log \: \frac{p^s(u)}{p^t(u)} \: du \Big{)} ds
\end{equation*}
\begin{equation*}
    = \int_s p(s) \Big{(} \int_u p^s(u) \: log \: \frac{E_{p(s)}[p^s(u)]}{p^s(u)} \: du + \: \int_u p^s(u) \: log \: \frac{p^s(u)}{p^t(u)} \: du \Big{)} ds
\end{equation*}
\begin{equation*}
    = - \: \E_{p(s)}[D_{KL}(p^s(u), E_{p(s)}[p^s(u)])] + \E_{p(s)}[D_{KL}(p^s(u), p^t(u))]
\end{equation*}
\begin{equation*}
    \leq \E_{p(s)}[D_{KL}(p^s(u), p^t(u))]
\end{equation*}

% =====================
% Variants of TTA-AE
% =====================
\subsection{TTA-AE variants}~\label{appendix_tta_ae_variants}
\cite{he2021autoencoder} propose a autoencoder-based method for TTA.
We made some minor changes in their method to get optimal results on the datasets used in our experiments. We did this analysis for 5 prostate segmentation test distributions, and used the optimal settings for the other datasets.

\vspace{5pt} \noindent \textbf{Architecture}:
% An important difference between the proposed work and \cite{he2021autoencoder} is in the parameters that are adapted for each test image.
In the proposed method, the adaptable module, $N_\phi$ is trained on the training distribution and further adapted for each test image.
In contrast, \cite{he2021autoencoder} introduce 4 \textit{adaptors}, $A^x$, $A^1$, $A^2$, $A^3$, as different layers in the task CNN directly at test time.
$A^1$, $A^2$, $A^3$ are initialized to be identity mappers, while $A^x$ is randomly initialized. 
In our experiments, we found that the randomly initialized $A^x$ (with the same architecture as in \cite{he2021autoencoder}) substantially altered the image intensities before any TTA iterations were done. Due to this, the Dice scores at the start of TTA iterations dropped to almost $0$, and could not be recovered by the TTA.
We could resolve this with the help of two changes to the architecture of $A^x$: (i) Instead of initializing the convolutional weights with mean $0$, we initialize with mean as the inverse of number input channels and variance as proposed in~\cite{he2015delving}, (ii) we removed instance normalization layers from $A^x$. The initial Dice scores (TTA epoch 0) were now reasonable ('Architecture' in Table \ref{tab:tta_ae_variants}), although much lower than the strong baseline. The TTA iterations improve the results, but are unable to cross the strong baseline.

\vspace{5pt} \noindent \textbf{Optimization}: We observed that the Dice scores fluctuated heavily across the TTA iterations. After reducing the learning rate from $0.001$ (used in \cite{he2021autoencoder}) to $0.00001$ and using the gradient accumulation strategy proposed in \cite{karani2020test}, we observed improved performance ('Optimization' in Table \ref{tab:tta_ae_variants}). However, the Dice scores initially improved and then dropped after about $100$ epochs, for $3$ of the $5$ test distributions.

\vspace{5pt} \noindent \textbf{Loss}: Plotting the evolution of the losses of the 5 AEs: one each at the input $AE^x$ and the output layers $AE^y$, and 3 at different features at different depths ($AE^{F1}$, $AE^{F2}$, $AE^{F3}$) in the task CNN, we observed that the accuracy of $AE^x$ and $AE^y$ correlated well with the Dice scores, while this was untrue for the feature-level AEs. Thus, we carried out TTA driven only by $AE^x$ and $AE^y$. In this setting, TTA-AE provided performance improvement in a stable manner ('Loss' in Table \ref{tab:tta_ae_variants}). We used this setting for the experiments on the rest of the datasets.

\begin{table}[h!]
    \centering
    \small
    \begin{tabular}{ c c c c c c }
        
        \hline
        \rowcolor{titlegray} 
        \backslashbox{Method}{Test}
        & UCL & HK & BIDMC & BMC & USZ \\ \hline
        
        \rowcolor{lightgray} & \multicolumn{5}{c}{{{Domain Generalization}}}
        \\ \hline
        \rowcolor{lightergray} 
        Strong baseline \cite{zhang2019generalizing} &
        0.77 & % UCL
        0.82 & % HK
        0.62 & % BIDMC
        0.78 & % BMC
        0.77 % USZ
        \\ \hline
        
        \rowcolor{lightgray} & \multicolumn{5}{c}{{{TTA-AE \cite{he2021autoencoder} Variants}}}
        \\ \hline
        
        % \rowcolor{lightgray} & \multicolumn{5}{c|}{{{Original hyper-parameters}}} \\\Xhline{2\arrayrulewidth}
        % \rowcolor{lightergray} TTA Epoch 0  & 0.08 & 0.01 & 0.02 & 0.09 & 0.01 \\ \hline
        % \rowcolor{lightergray} TTA Epoch 10  & 0.05 & 0.00 & 0.00 & 0.01 & 0.00 \\ \Xhline{2\arrayrulewidth}
        
        \rowcolor{lightgray} Modification in: & \multicolumn{5}{c}{{{Details}}} \\\hline
        
        \rowcolor{lightgray} Architecture & \multicolumn{5}{c}{{{Removing instance normalization in $A^x$}}} \\ \hline
        
        \rowcolor{lightergray} TTA Epoch 0
        & 0.76
        & 0.71
        & 0.48
        & 0.67
        & 0.57
        \\ % \hline
        
        \rowcolor{lightergray} TTA Epoch 10
        & 0.56
        & 0.73
        & 0.51
        & 0.50
        & 0.76
        \\ \hline
        
        \rowcolor{lightgray} Optimization & \multicolumn{5}{c}{{{Lower learning rate, gradient accumulation}}} \\ \hline
        
        \rowcolor{lightergray} TTA Epoch 0
        & 0.76
        & 0.71
        & 0.48
        & 0.67
        & 0.57
        \\ % \hline
        
        \rowcolor{lightergray} TTA Epoch 10
        & 0.78
        & 0.74
        & 0.50
        & 0.71
        & 0.65
        \\ % \hline
        
        \rowcolor{lightergray} TTA Epoch 100
        & 0.77
        & 0.83
        & 0.56
        & 0.78
        & 0.78
        \\ % \hline
        
        \rowcolor{lightergray} TTA Epoch 1000
        & 0.65
        & 0.78
        & 0.57
        & 0.73
        & 0.79
        \\ \hline
        
        \rowcolor{lightgray} Loss & \multicolumn{5}{c}{{{Using AEs only at input \& output layers}}} \\ \hline
        
        \rowcolor{lightergray} TTA Epoch 0
        & 0.76
        & 0.71
        & 0.48
        & 0.67
        & 0.57
        \\ % \hline
        
        \rowcolor{lightergray} TTA Epoch 10
        & 0.78
        & 0.74
        & 0.48
        & 0.71
        & 0.64
        \\ % \hline % 
        
        \rowcolor{lightergray} TTA Epoch 100
        & 0.79
        & 0.82
        & 0.51
        & 0.78
        & 0.78
        \\ % \hline
        
        \rowcolor{lightergray} TTA Epoch 1000
        & 0.78
        & 0.83
        & 0.50
        & 0.79
        & 0.79
        \\ \hline
        
        % \rowcolor{lightgray} & \multicolumn{5}{c|}{{{Adapting $N_\phi$ instead of $A^x$, $A^1$, $A^2$, $A^3$}}} \\\Xhline{2\arrayrulewidth}
        % \rowcolor{lightergray} TTA Epoch 0 &  0. & 0. & 0. & 0. & 0. \\ \hline
        % \rowcolor{lightergray} TTA Epoch 10 &  0. & 0. & 0. & 0. & 0. \\ \hline
        % \rowcolor{lightergray} TTA Epoch 100 &  0. & 0. & 0. & 0. & 0. \\ \hline
        % \rowcolor{lightergray} TTA Epoch 1000 &  0. & 0. & 0. & 0. & 0. \\ \Xhline{4\arrayrulewidth}

    \end{tabular}
    \caption[TTA-AE - Hyper-parameter tuning]{Performance of TTA-AE~\cite{he2021autoencoder} variants.}~\label{tab:tta_ae_variants}
    \vspace{-25pt}
\end{table}

\end{document}